\documentclass{article}
\usepackage{caption}  
\usepackage[accepted]{icml2025}

\icmltitlerunning{Adaptive Sample Sharing for Multi Agent Linear Bandits}

\usepackage[utf8]{inputenc}                      
\usepackage{natbib}                              
\usepackage[T1]{fontenc}                         
\usepackage{hyperref}                            
\usepackage{url}                                 
\usepackage{booktabs}                            
\usepackage{amsthm,amsmath,amssymb,amsfonts}     
\usepackage{dsfont}                              %
\usepackage{thmtools,thm-restate}                
\usepackage{algorithm,algorithmic}               
\usepackage{mdframed}                            
\usepackage[many]{tcolorbox}                     
\usepackage{tabularx}                            
\usepackage{nicefrac}                            
\usepackage{microtype}                           
\usepackage{xcolor}                              
\usepackage{bm}                                  
\usepackage{graphicx}                            
\usepackage{enumitem}                            
\usepackage{aliascnt}                            
\usepackage{sidecap}                             
\usepackage{makecell}                            
\usepackage{array}                               
\usepackage{subcaption}                          
\usepackage{dashrule }                           

\declaretheorem[name=Lemma,numberwithin=section]{lemma}

\declaretheorem[name=Definition,numberwithin=section]{definition}
\declaretheorem[name=Assumption,numberwithin=section]{assumption}

\DeclareMathOperator*{\argmax}{arg\,max}

\def\Nithat{\widehat{\mathcal{N}}_{i_t, t-1}}

\def\ANithat{\bm{A}_{\widehat{\mathcal{N}}_{i_t, t-1}}}

\def\thetaNithat{\widehat{\bm{\theta}}_{\widehat{\mathcal{N}}_{i_t, t-1}}}

\def\eg{\emph{e.g.},~}
\def\ie{\emph{i.e.},~}
\def\wrt{w.r.t.~}
\def\resp{resp.~}
\def\orangetri{\begingroup\color{tab:orange}$\blacktriangle$\endgroup}
\def\bluetri{\begingroup\color{tab:blue}$\blacktriangle$\endgroup}
\def\orangecirc{\begingroup\color{tab:orange}$\bullet$\endgroup}
\def\bluecirc{\begingroup\color{tab:blue}$\bullet$\endgroup}
\newcommand{\reddashedline}{\textcolor{red}{\hdashrule[0.5ex]{1cm}{1pt}{3pt}}}

\def\equationautorefname~#1\null{(#1)\null}

\definecolor{tab:blue}{RGB}{31, 119, 180}
\definecolor{tab:orange}{RGB}{255, 127, 14}
\definecolor{linkcolor}{RGB}{44, 96, 163}
\hypersetup{
    colorlinks=true,
    citecolor=linkcolor,
    linkcolor=linkcolor,
    urlcolor=linkcolor,
}

\begin{document}

\twocolumn[
\icmltitle{Adaptive Sample Sharing for Multi Agent Linear Bandits}

\icmlsetsymbol{equal}{*}

\begin{icmlauthorlist}
\icmlauthor{Hamza Cherkaoui}{a}
\icmlauthor{Merwan Barlier}{a}
\icmlauthor{Igor Colin}{b}
\end{icmlauthorlist}

\icmlaffiliation{a}{Huawei Noah’s Ark Lab, France}
\icmlaffiliation{b}{LTCI, T\'el\'ecom Paris, Institut Polytechnique de Paris, France}
\icmlcorrespondingauthor{Hamza Cherkaoui}{hamza.cherkaoui@huawei.com}
\icmlkeywords{Linear bandit, Collaboration, Sample sharing}

\vskip 0.3in
]

\printAffiliationsAndNotice{}

\begin{abstract}
The multi-agent linear bandit setting is a well-known setting for which designing efficient collaboration between agents remains challenging.
This paper studies the impact of data sharing among agents on regret minimization.
Unlike most existing approaches, our contribution does not rely on any assumptions on the bandit parameters structure.
Our main result formalizes the trade-off between the bias and uncertainty of the bandit parameter estimation for efficient collaboration.
This result is the cornerstone of the Bandit Adaptive Sample Sharing (BASS) algorithm, whose efficiency over the current state-of-the-art is validated through both theoretical analysis and empirical evaluations on both synthetic and real-world datasets.
Furthermore, we demonstrate that, when agents' parameters display a cluster structure, our algorithm accurately recovers them.
\end{abstract}

\section{Introduction}
\label{sec:intro}

The stochastic multi-armed bandit (MAB) framework provides efficient tools for solving sequential optimization problems~\citep{Lai1985, Abbasi2011, dani2008stochastic, soare2015sequential}. In this setting, an agent repeatedly pulls an arm from a finite set of candidates and observes the associated noisy reward, sampled from an unknown fixed distribution.

A common extension of the MAB problem is the linear bandit setting~\citep{tor2020}, where each arm is associated with a vector $\bm{x} \in \mathbb{R}^d$ and its expected reward is a linear combination of the vector and an unknown bandit parameter $\bm{\theta}^* \in \mathbb{R}^d$. The linear bandits framework is a generalization of MAB that takes into account the similarities between the arms to identify better pulling strategies. Efficiently solving this problem often consists in obtaining an accurate estimate of the parameter $\bm{\theta}^*$ with respect to some metric. For example, in the context of regret minimization~\citep{Li2010, Chu2011}, the goal is to obtain an estimate of $\bm{\theta}^*$ that is particularly accurate in the directions associated with large rewards.

In the last decade, there has been an increasing focus on the multi-agents setting~\citep{Bianchi2013, Gentile2014, Nguyen2014, Ban2021, Ghosh2022, Do2023, Yang2024}, where multiple agents share samples to make better decisions together. For example, consider the problem of optimizing the parameters of wireless antennas in a network to improve the quality of service for nearby users. Antennas that belong to similar environments and constraints should yield similar rewards and are likely to benefit from sharing their observations, at least in the early stages of the optimization process. Indeed, if the rewards are derived from a similar parameter $\bm{\theta}^*$, sharing observations effectively reduces the noise variance, making the estimator more accurate. Similar arguments can be made for recommendation systems or drug trials~\citep{sarwar2002incremental, su2009survey, Li2010, Chu2011}, where similar profiles should lead to similar observations.

This problem, known in the literature as Heterogeneous Multi-Agent Linear Stochastic Bandits, has been mostly approached using the Euclidean similarity to define agents' clusters~\citep{Gentile2014, Nguyen2014} and allow sample sharing within them. 

This paper develops a new approach to determine, \emph{without assumption on a clustered structure}, when agents should share samples, focusing on the bias introduced by heterogeneous agents, the resulting reduction in uncertainty, and the trade-offs between these factors. Based on our similarity metric on the Mahalanobis distance induced by the observations to focus on \emph{directions with large reward} and avoiding using the overly conservative Euclidean distance, we introduce the Bandit Adaptive Sample Sharing (BASS) algorithm. This method outperforms the state-of-the-art by pairing similarity learning with regret minimization.

We organize our paper as follows: first, in~\autoref{sec:related_work}, we detail how this problem was addressed in the literature and summarize our contributions, \autoref{sec:preliminaries} provides the intuition on how our approach challenges it. Then in~\autoref{sec:single_case}, we formalize the linear bandit setting for the single agent case. \autoref{sec:collaboration_case} explores sample sharing in a multi-agent context and introduces our sharing criterion. In~\autoref{sec:bass_and_analysis}, we present and detail our algorithm, followed by a theoretical analysis. Finally, we empirically evaluate our method in~\autoref{sec:exp}.

\section{Related Work}
\label{sec:related_work}

\paragraph{Cluster structure known in advance} \quad Some work has assumed that the parameter structure is known in advance. \citet{Bianchi2013} modifies the problem variables to consider all agents simultaneously, allowing them to mimic a classical single-agent scenario while proposing an efficient multi-agent algorithm. In \citet{Wu2016}, the problem is modeled to compute each reward as a linear combination of neighbors parameters, leading to a collaborative setting. More recently, \citet{Moradipari2022} defined a setting in which agents aim to maximize the average reward of the network, forcing agents to share their observations during a communication phase.

In real-world scenarios, however, the parameter structure is often unknown, so a natural goal is to recover it before enabling collaboration.

\paragraph{Estimating the cluster structure as a graph}\quad The seminal work of~\citet{Gentile2014} introduces a novel collaborative framework: agents are grouped into clusters, and agents within the same cluster have the Euclidean distance between their bandit parameters smaller than a given threshold. With this assumption, the authors derive a similarity measure based on this distance and use the corresponding graph to recover clusters. Since then, variants have been proposed to improve the accuracy of the parameter estimates, \eg by averaging over connected components rather than neighborhoods~\citep{Li2018}, or by improving the robustness of the estimates by offsetting the OLS~\citep{Wang2023}
. Other work has focused on improving the quality of the clustering itself: \citet{Ban2021} allows overlapping clusters, \citet{Li2019, Xu2024} each proposes a splitting and merging procedure to correct for potential clustering errors, and~\citet{Ghosh2022} aims to control the minimum cluster size.
Although still based on a Euclidean similarity measure, \citet{cheng23} opts for a different clustering routine to take advantage of hedonic game theory. \citet{Do2023} and~\citet{Yang2024} trigger collaboration rounds, controlled by a collaboration budget, when the design matrices deviate too much from each other. Finally,~\citet{Li2016, Gentile2017} base their similarity on the rewards themselves, rather than the parameters, alleviating some of the dimensionality issues of the Euclidean similarity.

While most contributions have relied or focused on improving the clustering process with the Euclidean similarity introduced by~\citet{Gentile2014}, this paper takes advantage of the Mahalanobis distance, as regret minimization problems require focusing primarily on \emph{on the directions with the highest rewards}. Furthermore, unlike most existing approaches that rely on the assumption of an \emph{agent clustered structure}, our contribution does not impose such constraint.

\paragraph{Multivariate approaches} \quad In an opposite research direction, the approach in~\citet{Nguyen2014} takes advantage of the $k$-means~(\citet{MacQueen1967}) algorithm to cluster the agents. Alternatively, instead of making the hypothesis of clusters, other approaches propose to enforce some regularization on the bandit parameter estimation to introduce structure between agents. The Non-negative Matrix Factorization (NMF) technique in~\citet{Le2018} allows them to recover a probability distribution on the cluster labels for each agent. Finally, a low-rank decomposition of the stacked bandit parameter matrix in ~\citet{Yang2020} yields the desired cluster structure of the problem, by combining the decomposition with a sparsity constraint. 

However, the multivariate aspect of these approaches makes them less suitable for a distributed framework, which is often required in real-world applications (\eg{} recommendation systems, wireless antenna network).

\subsection{Our contributions}

We propose a novel algorithm that addresses several previous limitations and introduces key features that distinguish it from previous work.

\begin{enumerate}[itemsep=0.25pt, topsep=0.25pt]
    \item \emph{No Assumption on Bandit Parameters}: Our approach does not rely on any assumptions on the bandit parameter structure to enable collaboration, and it learns to balance the trade-off between the collaborative bias and the uncertainty reduction.
    \item \emph{Anisotropic Approach}: In regret minimization, we focus on directions with large reward; our approach therefore focuses on the same directions to quantify the bias with adequate precision.
    \item \emph{Theoretical and Empirical Analysis}: We formally define the problem, conduct an exhaustive analysis of collaborative stopping time, and cumulative and instantaneous regret. In addition, we present a fair and comprehensive empirical evaluation of our algorithm using both synthetic and real-world data.
\end{enumerate}

\section{Preliminaries}
\label{sec:preliminaries}

\subsection{Notation}

We use lowercase (\eg $\alpha$) to denote a scalar, bold (\eg $\bm{x}$) to denote a vector, and uppercase bold (\eg $\bm{A}$) is reserved for a matrix. The $\ell_2$-norm of a vector $\bm{x}$ is $\|\bm{x}\|_2 = \sqrt{\bm{x}^\top \bm{x}}$ and the Mahalanobis weighted $\ell_2$-seminorm is $\|\bm{x}\|_{\bm{A}} = \sqrt{\bm{x}^\top \bm{A} \bm{x}}$, where $\bm{A} \succeq 0$ is semidefinite positive. The ellipsoid of center $\bm{c} \in \mathbb{R}^d$, shape $\bm{A} \succeq 0$ and radius $r$ is denoted $\mathcal{E}(\bm{c}, \bm{A}, r)$, that is
\[
    \mathcal{E}(\bm{c}, \bm{A}, r) = \left\{ \bm{x} \in \mathbb{R}^d \text{, } \|\bm{x} - \bm{c} \|_{\bm{A}} \leq r \right\} \enspace.
\]

An exhaustive notation summary can be found in the supplementary material~\autoref{sec:supp_notations}.

\subsection{A simple two agent problem}
\label{subsec:main_idea}

To build intuition about sample sharing in the regret minimization problem, we first examine the simpler case of parameter estimation with two agents.

Consider two agents, each one with the aim of estimating its linear parameter $\bm{\theta}^*_i \in \mathbb{R}^d$, $i \in \{1, 2\}$. To do this, each agent $i$ has a series of $t$ actions $(\bm{x}_{i, 1}, \ldots, \bm{x}_{i, t})$ associated with a series of $t$ noisy rewards $(y_{i, 1}, \ldots, y_{i, t})$ such that for $1 \leq s \leq t$:
\[
    y_{i, s} = {\bm{\theta}^*_i}^\top \bm{x}_{i, s} + \eta_{i, s} \enspace,
\]
for some $1$ sub-Gaussian\footnote{see~\autoref{assum:sub_gaussian_noise} for a formal definition of sub-Gaussianity.} noise $\eta_{i, s}$. The unbiased Ordinary Least Squares (OLS) estimator is then:
\[
    \hat{\bm{\theta}}_{i, t} = \left( \sum_{s = 1}^t \bm{x}_{i, s} \bm{x}_{i, s}^\top \right)^+ \sum_{s = 1}^t y_{i, s} \bm{x}_{i, s} = \bm{\theta}^*_i + \bm{A}_{i, t}^+ \bm{z}_{i, t} \enspace,
\]
where $\bm{A}_{i, t} = \sum_{s = 1}^t \bm{x}_{i, s} \bm{x}_{i, s}^\top$, $\bm{z}_{i, t} = \sum_{s = 1}^t \eta_{i, s} \bm{x}_{i, s}$ and for any $\bm{M} \in \mathbb{R}^{d \times d}$, $\bm{M}^+$ is the Moore-Penrose inverse. For simplicity, we assume that both agents hold the same series of actions, so $\bm{A}_{1, t} = \bm{A}_{2, t} = \bm{A}_t$.

If the two agents shared their observations, the OLS of the merged dataset would be
\[
    \hat{\bm{\theta}}_{\mathrm{c}, t} = \bm{\theta}^*_\mathrm{c} + \frac{1}{2} \bm{A}_t^+ (\bm{z}_{1, t} + \bm{z}_{2, t})\enspace,
\]
where $\bm{\theta}^*_\mathrm{c} = \frac{1}{2}(\bm{\theta}^*_1 + \bm{\theta}^*_2)$. If both parameters are equal, then the merged OLS improves each independent OLS, since it is unbiased and the noise is only $\frac{1}{\sqrt{2}}$ sub-Gaussian. However, if the parameters are different, the merged OLS is biased, and the question becomes whether the noise reduction outweighs the bias.

Since the overall goal is to consider the regret minimization problem, we measure the accuracy of an estimator by its Mahalanobis distance w.r.t.\ the design matrix $\bm{A}_t$. Indeed, the larger the reward, the more accurate the estimate needs to be. Using the analysis of~\citet{Abbasi2011}, we have, with high probability,
\[
    \| \hat{\bm{\theta}}_{i, t} - \bm{\theta}^*_i \|_{\bm{A}_t} \leq \beta_t \enspace,
\]
where $\beta_t = O(\sqrt{\log(t)})$ will be detailed later. Similarly,
\[
    \|\hat{\bm{\theta}}_{\mathrm{c}, t} - \bm{\theta}_i^{*}\|_{\bm{A}_t} \leq \frac{1}{2} \| \bm{\theta}_1^{*} -  \bm{\theta}_2^{*}\|_{\bm{A}_t} + \frac{1}{\sqrt{2}} \beta_t \enspace.
\]
Therefore, the collaborative estimate $\hat{\bm{\theta}}_{\mathrm{c}, t}$ improves the regular OLS estimator when
\begin{equation}
    \label{eq:sep_crit}
    \| \bm{\theta}_1^{*} -  \bm{\theta}_2^{*}\|_{\bm{A}_t} \leq (2 - \sqrt{2}) \beta_t \enspace.
\end{equation}

Equation~\autoref{eq:sep_crit} suggests that collaboration reduces the error as long as the bias introduced $\| \bm{\theta}_1^{*} -  \bm{\theta}_2^{*}\|_{\bm{A}_t}$ is less than a fraction of the uncertainty of the estimation $\beta_t$. 

With this simple case, we argue that, given a sampling approach of the arms $(\bm{x}_{s})_{s=1}^t$, we need to adapt \emph{when the collaboration should occur} to accelerate the estimation process. The key point is to detect when the condition of Equation~\autoref{eq:sep_crit} stops being verified to end the collaboration and to avoid introducing a detrimental bias.

\begin{figure}[ht]
    \centering
    \begin{minipage}{0.24\textwidth}
        \centering
        \includegraphics[width=.8\linewidth,trim= 15 40 25 30,clip]{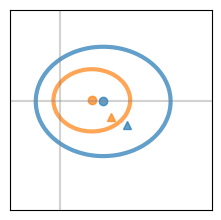}
    \end{minipage}%
    \hfill
    \begin{minipage}{0.24\textwidth}
        \centering
        \caption{Estimation of $\bm{\theta}_1^{*}$ depicted with \bluecirc{} (\resp{} $\bm{\theta}_{\mathrm{c}}^{*}$ with \orangecirc{}), $\hat{\bm{\theta}}_{1, t}$ with \bluetri{} (\resp{} $\hat{\bm{\theta}}_{\mathrm{c}, t}$ with \orangetri{}), along with the corresponding confidence ellipsoid in \textcolor{tab:blue}{blue} (\resp{} in \textcolor{tab:orange}{orange}). The collaborative estimate has a reduced uncertainty ellipsoid $\|\hat{\bm{\theta}}_{\mathrm{c}, t} - \bm{\theta}_i^{*}\|_{\bm{A}_t}$.}
        \label{fig:illustration}
    \end{minipage}
\end{figure}

The later intuitive result can be easily illustrated. In~\autoref{fig:illustration}, we display $\bm{\theta}_1^{*}$ and $\hat{\bm{\theta}}_t$ along with their confidence ellipsoid in blue and their collaborative counterparts in orange. We notice the well-known estimation bias - variance trade-off.

\section{Single agent regret minimization}
\label{sec:single_case}

In the linear bandit setting, at each time step $t$ an agent chooses an arm $\bm{x}_t$ from a set $\mathcal{X} = (\bm{x}_k)_{k=1}^K \in \mathbb{R}^d$ and receives a reward signal $y_t= \bm{x}_t^\top \bm{\theta}^* + \eta$, where $\bm{\theta}^*$ is the (unknown) bandit parameter and $\eta$ is some centered noise. For the remainder of the paper, we assume that the arms and the bandit parameters are bounded as depicted in~\autoref{assum:bounded_norms}

\begin{assumption}[Bounded Norms]
\label{assum:bounded_norms}

We assume there exists a constant \( L > 0 \) such that the true parameter satisfies \( \| \bm{\theta}^* \| \leq L \), and all context vectors satisfy \( \| \bm{x}_k \| \leq 1 \) for all \( 1 \leq k \leq K \).

\end{assumption}

Let $\mathcal{F}_t$ be the $\sigma$-algebra generated by $(\bm{x}_1, y_1, \ldots, \bm{x}_{t-1}, y_{t-1}, \bm{x}_{t})$, then $y_{t}$ is {$\mathcal{F}_{t-1}$-measurable}. Moreover, we make the common~\autoref{assum:sub_gaussian_noise} that the noise is conditionally $R$-sub-Gaussian for some constant $R>0$. 

\begin{assumption}[$R$-sub-Gaussianity of the Noise]
\label{assum:sub_gaussian_noise}

The noise term \( \eta \) is conditionally \( R \)-sub-Gaussian for some constant \( R > 0 \). That is, for all \( \lambda \in \mathbb{R} \),
\[
    \mathbb{E}\left[ \exp(\lambda \eta) \mid \mathcal{F}_{t-1} \right] \leq \exp\left( \frac{\lambda^2 R^2}{2} \right) \enspace.
\]

\end{assumption}

The goal is to minimize the cumulative pseudo-regret, formally defined in~\citet{Audibert2009} as $R_T = \sum_{t=1}^T r_t \quad \text{, with} \quad r_t= \bm{\theta}^{* \top} (\bm{x}^* - \bm{x}_{t})$ and $\bm{x}^* = \argmax_k \bm{\theta}^{* \top} \bm{x}_k$. For this purpose, the OFUL algorithm~\citep{Abbasi2011} implements the principle of optimism in the face of uncertainty in the linear bandit case. At each time step $t$, the learner pulls the arm that maximizes the expected reward associated with the best parameter in a confidence region around the Ordinary Least Squares (OLS) estimator. Formally, at time $t + 1$, given a previous OLS estimator $\hat{\bm{\theta}}_t$ and a $\mathcal{C}_\delta(\hat{\bm{\theta}}_{t})$, the selected arm maximizes the optimist reward
\[
    \bm{x}_{t + 1} \in \argmax_{\bm{x} \in \mathcal{X}} \max_{\bm{\theta} \in \mathcal{C}_\delta(\hat{\bm{\theta}}_t)} \bm{x}^\top \bm{\theta} \enspace,
\]
where $\mathcal{C}_\delta(\hat{\bm{\theta}}_t)$ is such that $\mathbb{P}(\bm{\theta}^* \in \mathcal{C}_\delta(\hat{\bm{\theta}}_t)) \geq 1 - \delta$. In practice, the confidence ellipsoid $\mathcal{C}_\delta(\hat{\bm{\theta}}_t)$ is constructed from the previous pulls using Hoeffding's concentration inequality~\citep{tropp2015introduction}, which leads to the theorem~\autoref{th:RHS_est_theta} from~\citet{Abbasi2011}:

\begin{restatable}[Confidence Ellipsoid for Bandit Parameter Estimation]{theorem}{thma}
\label{th:RHS_est_theta}

Let \( \delta \in (0, 1) \), \( t > 0 \), and \( \bm{\theta}^* \in \mathbb{R}^d \). Let \( (\bm{x}_s)_{1 \leq s \leq t} \) denote the sequence of arms pulled up to time \( t \). Under \autoref{assum:bounded_norms} and \autoref{assum:sub_gaussian_noise}, the following holds with probability at least \( 1 - \delta \):
\[
    \mathbb{P} \left( \bm{\theta}^* \in \mathcal{C}_\delta(\hat{\bm{\theta}}_t) \right) \geq 1 - \delta \enspace,
\]
where
\begin{equation}
    \label{eq:confidence_ellipsoid}
    \begin{cases}
        \mathcal{C}_\delta(\hat{\bm{\theta}}_t) = \left\{ \left\| \hat{\bm{\theta}}_t - \bm{\theta}^* \right\|_{\bm{A}_t} \leq \beta(\delta, \bm{A}_t) \right\} \enspace, \\
        \beta(\delta, \bm{A}_t) = R \sqrt{2 \log\left( \dfrac{1}{\delta} \sqrt{\dfrac{\det(\bm{A}_t)}{\det(\bm{A}_0)}} \right)} \enspace,
    \end{cases}
\end{equation}
and \( \bm{A}_t = \sum_{s = 1}^{t} \bm{x}_s \bm{x}_s^\top \) is the empirical design matrix.
\end{restatable}

 For the remainder of the paper, we assume that $\mathcal{X}$ spans $\mathbb{R}^d$. Also, we set $\bm{A}_{0} = \bm{I}$, so:
 \[
 \beta(\delta, \bm{A}_t) = R \sqrt{2 \log \frac{1}{\delta} +  \log \mathrm{det}(\bm{A}_{t}) } \enspace.
 \]
 
\section{Adaptive sample sharing}
\label{sec:collaboration_case}

In this section, we consider the collaborative setting and examine the impact of sharing samples on the regret minimization.

\subsection{Collaborative setting}

We consider a multi-agent setting where each agent $i \in \{1, \ldots, N\}$ is able to pull arms from a linear bandit of unknown parameters $\bm{\theta}^*_i \in \mathbb{R}^d$. Every linear bandit uses the same set of arms $\mathcal{X}$. We assume that the historical data are known to a central controller. At each time step $t$, the controller selects the arm to be pulled by the agents and observes the noisy rewards.

We are interested in comparing two pulling strategies: a local strategy, where each agent can only use its own observations to derive a pulling policy, and a collaborative strategy, where agents are allowed to regroup their observations in order to derive more general policies.

Throughout the paper, we denote by $\widehat{\mathcal{N}}_i(t)$ the set of agents sharing their data with agent $i$ at time $t$, we index a quantity by $i$ if it is computed using only local observations of agent $i$, and by $\widehat{\mathcal{N}}_i(t)$ if it is computed using observations from the set of all agents that share their data with $i$ at time $t$. In both cases, we focus on an OFUL approach, so the two strategies can be summarized as follows.

\textbf{Local strategy} use OFUL with local observations, compute the local OLS $\hat{\bm{\theta}}_i$ and the confidence region $\mathcal{C}_\delta(\hat{\bm{\theta}}_i)$;

\textbf{Collaborative strategy} use OFUL with the observations of the collaborating agents, compute the associated OLS $\hat{\bm{\theta}}_{\widehat{\mathcal{N}}_i(t)}$ and the confidence region $\mathcal{C}_\delta(\hat{\bm{\theta}}_{\widehat{\mathcal{N}}_i(t)})$.

\subsection{Separation test}
\label{subsec:regret_upper_bounds}

We now analyze the impact of sample sharing on the regret and formally introduce the test to detect when beneficial sharing stops.

The arm selection strategy is based on OFUL and depends on whether or not the observations are shared to define the OLS estimator and the confidence ellipsoid. Unlike the analysis in~\autoref{subsec:main_idea}, we focus on regret minimization rather than parameter estimation. To this end, we recall the following bounds on OFUL instantaneous regret for both independent and data-sharing settings.

\begin{restatable}[Instantaneous Regret Upper Bounds]{lemma}{lema}
\label{lem:regret_bound_inst}

Let \( 0 < \delta < 1 \) and \( t > 0 \). Let \( (\bm{\theta}^*_j)_{1 \leq j \leq N} \in \mathbb{R}^{d \times N} \) be the true parameters of \( N \) linear bandit models. Denote by \( \bm{x}_{i,t} \) (resp. \( \bm{x}_{\widehat{\mathcal{N}}_i(t)} \)) the arm selected by agent \( i \) using the local (resp. collaborative) strategy.

Then, with probability at least \( 1 - \delta \), the instantaneous regret of agent \( i \) is bounded as follows:

\begin{enumerate}[label=(\roman*)]
    \item \textbf{Local strategy}:
    \[
        r_i(t) \leq 2 \beta\left(\delta, \bm{A}_{i,t}\right) \left\| \bm{x}_{i,t} \right\|_{\bm{A}_{i,t}^{-1}} \enspace.
    \]

    \item \textbf{Collaborative strategy}:
    \[
        r_i(t) \leq \frac{2 \beta(\delta, m \bm{A}_{i,t})}{\sqrt{m}} \left\| \bm{x}_{\widehat{\mathcal{N}}_i(t)} \right\|_{\bm{A}_{i,t}^{-1}} + \frac{2}{m} \sum_{j \in \widehat{\mathcal{N}}_i(t)} \Delta^{i,j}_{\bm{A}_{i,t}} \enspace,
    \]
\end{enumerate}
where \( m = |\widehat{\mathcal{N}}_i(t)| \) and \( \Delta^{i,j}_{\bm{A}_{i,t}} = \left\| \bm{\theta}^*_i - \bm{\theta}^*_j \right\|_{\bm{A}_{i,t}} \).
\end{restatable}

Following the analysis of~\autoref{subsec:main_idea} and substituting the unknown parameters in their OLS estimates, we compare the right-hand side of the inequalities. For $j \in \widehat{\mathcal{N}}_i(t)$, we outline the same condition than Equation~\autoref{eq:sep_crit}, see~\autoref{sec:supp_initial_sep_criteria}: 
\begin{equation}
    \label{eq:stop_crit}
    \| \hat{\bm{\theta}}_i - \hat{\bm{\theta}}_j \|_{\bm{A}_{i,t}} \leq (2 - \sqrt{2}) \beta(\delta, \bm{A}_t) \enspace.
\end{equation}

We almost recover the overlapping ellipsoid test derived in~\citet{Gilitschenski2012} and based on the following function:
\begin{equation}
    \label{eq:gili_func}
    \begin{split}
        \kappa(i, j) = 1 - \underset{s \in ]0,1[}{\min} \| \hat{\bm{\theta}}_i - \hat{\bm{\theta}}_j \|_{\left(\frac{\beta_i}{1-s} \bm{A}_{i,t}^{-1} + \frac{\beta_k}{s} \bm{A}_{j,t}^{-1} \right)^{-1}} \enspace.
    \end{split}
\end{equation}
The ellipsoid separation property is obtained from the sign of $\kappa(i, j)$, that is $\mathcal{E}(\hat{\bm{\theta}}_i, \bm{A}_{i,t}, \beta_i) \cap \mathcal{E}(\hat{\bm{\theta}}_j, \bm{A}_{j,t}, \beta_j) = \emptyset$ if and only if $\kappa(i, j) \leq 0$.

Since we consider a synchronous pulling, we have $\bm{A}_{i,t} = \bm{A}_{j,t} = \bm{A}_t$, and $\beta_i = \beta_j = \beta(\delta, \bm{A}_t)$. Indeed, in this setting, all collaborating agents select an arm based on the same shared history. Choosing a deterministic arm policy yields that two agents using the same history will select the same arm. Therefore, their design matrices will also be identical. The separation condition can be reformulated as stated in the following corollary.

Since we consider a synchronous setting, we have \( \bm{A}_{i,t} = \bm{A}_{j,t} = \bm{A}_t \) and \( \beta_i = \beta_j = \beta(\delta, \bm{A}_t) \). In this case, all collaborating agents operate on the same shared history. Because the arm selection policy (\eg{} UCB) is deterministic, agents with identical histories will always select the same arms. As a result, their design matrices evolve identically over time.

Under this assumption, the separation condition can be reformulated as stated in the following corollary.

\begin{restatable}[Ellipsoid Separation Under Synchronous Pulling]{lemma}{lemb}
\label{cor:ell_test}

Let \( 0 < \delta < 1 \), and consider two agents \( i \) and \( j \). Under \autoref{assum:bounded_norms} and \autoref{assum:sub_gaussian_noise}, define \( \tilde{\beta} = (1 + 1/\sqrt{2}) \beta(\delta, \bm{A}_t) \). Then the following statements are equivalent:
\begin{enumerate}[label=(\roman*)]
    \item \( \left\lVert \hat{\bm{\theta}}_i - \hat{\bm{\theta}}_j \right\rVert_{\bm{A}_t} \geq (2 + \sqrt{2}) \beta(\delta, \bm{A}_t) \enspace, \)
    \item \( \mathcal{E}(\hat{\bm{\theta}}_i, \bm{A}_t, \tilde{\beta}) \cap \mathcal{E}(\hat{\bm{\theta}}_j, \bm{A}_t, \tilde{\beta}) = \emptyset \enspace. \)
\end{enumerate}
\end{restatable}

\begin{figure}[ht]
    \centering
    \begin{minipage}{0.25\textwidth}
        \centering
        \includegraphics[width=.9\linewidth,trim= 10 70 10 20,,clip]{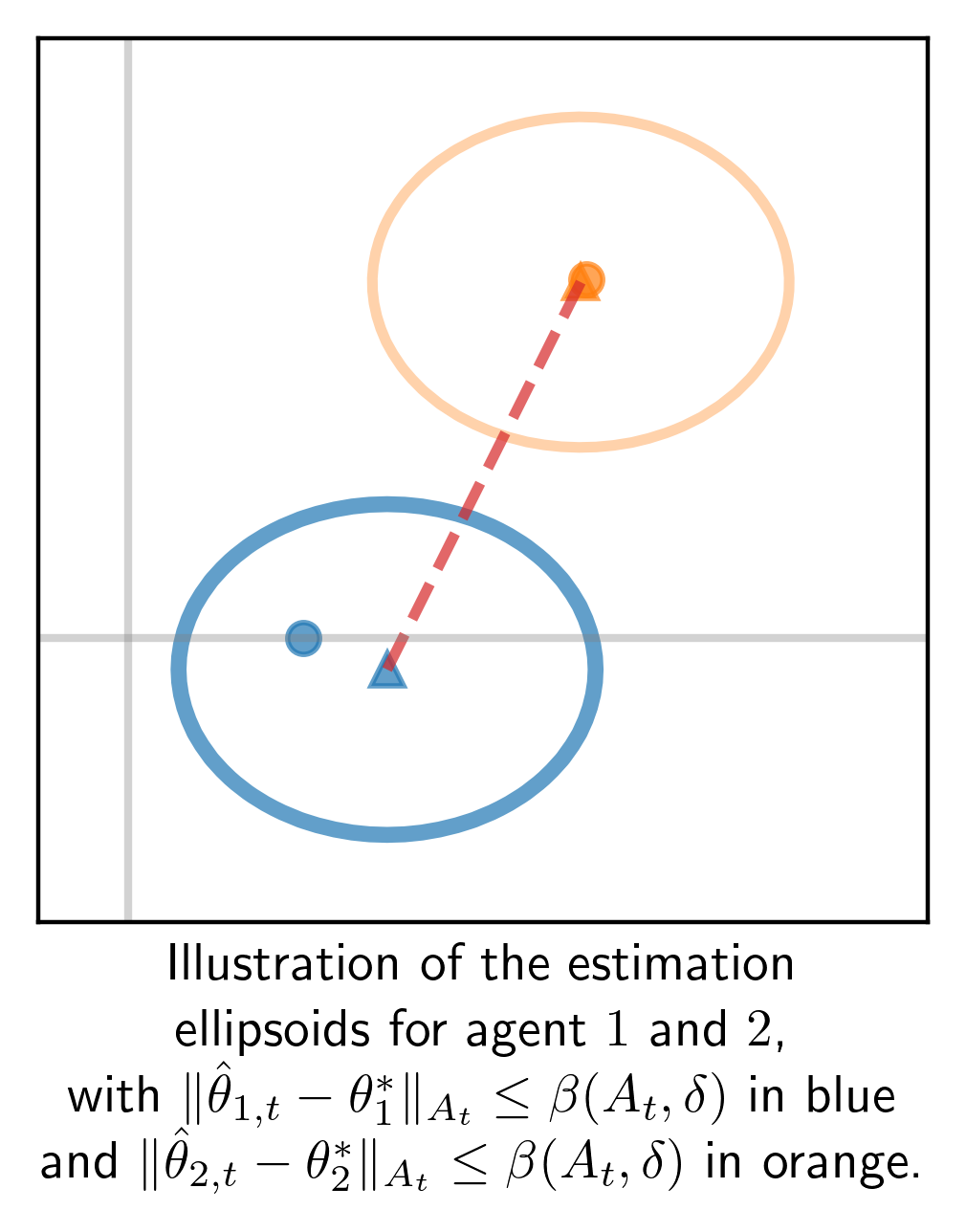}
    \end{minipage}%
    \hfill
    \begin{minipage}{0.23\textwidth}
        \centering
        \caption{Estimation of $\bm{\theta}_i^{*}$ depicted with \bluecirc{} (\resp{} $\bm{\theta}_j^{*}$ with \orangecirc{}), $\hat{\bm{\theta}}_{i, t}$ with \bluetri{} (\resp{} $\hat{\bm{\theta}}_{j, t}$ with \orangetri{}), along with the corresponding confidence ellipsoid in \textcolor{tab:blue}{blue} (\resp{} in \textcolor{tab:orange}{orange}). With high probability, the bias, depicted with a red dashed line~\reddashedline{}, becomes detrimental when the ellipsoids are separated.}
        \label{fig:illustration_ellipsoids}
    \end{minipage}
\end{figure}

This shed a new light on the preliminary work in~\autoref{subsec:main_idea} and~\autoref{subsec:regret_upper_bounds}. The collaborative condition in Equation~\autoref{eq:sep_crit} matches the separation condition for ellipsoids $\mathcal{E}(\hat{\bm{\theta}}_i, \bm{A}_t, \tilde{\beta})$ and $\mathcal{E}(\hat{\bm{\theta}}_j, \bm{A}_t, \tilde{\beta})$.
Interestingly, denoting as $\tilde{\delta}$ the value for which $\tilde{\beta} = \beta(\tilde{\delta}, \bm{A}_t)$, the bias is detrimental when we know with probability at least $1 - \tilde{\delta}$ that both agents follow two different bandit parameters, as illustrated in~\autoref{fig:illustration_ellipsoids}.

It should be noted that collaboration criteria based on the Euclidean distance (\eg{}~\citet{Gentile2014,Ban2021,Wang2023}) are equivalent to the~\citet{Gilitschenski2012} test where the ellipsoids considered are $\mathcal{E}(\hat{\bm{\theta}}_i, \lambda_{\min}(\bm{A}_t)\bm{I}, \tilde{\beta})$ and $\mathcal{E}(\hat{\bm{\theta}}_j, \lambda_{\min}(\bm{A}_t)\bm{I}, \tilde{\beta})$, \emph{ignoring most of the arm pulling history}.

In practice, computing~\eqref{eq:gili_func} is expensive, instead we consider a cheaper, equivalent test function\footnote{We proof the ellipsoid separation property of $\Psi$ in the supplementary material~\autoref{sec:supp_our_test_equivalence}}, defined as follows.

\begin{definition}[$\gamma$-relaxed ellipsoid separation test function]
    \label{def:ell_test}

    For any iteration $t$ and any two agents $i$ and $j$, we denote as $\Psi(i, j, t)$ the quantity
    \[
        \Psi(i, j, t) = \mathds{1} \left\{ \min_{s \in [0,1]} \tau_t^{i,j}(s) < 0 \right\} \enspace,
    \]
    with
    \[
        \tau_t^{i,j}(s) = \frac{\gamma^2}{4} - \sum_{l=1}^d \mu_{t,l}^2 \cdot \frac{s(1 - s)}{\beta_{i,t}^2 + s \left( \beta_{j,t}^2 \eta_{t,l} - \beta_{i,t}^2 \right)}
    \]
    where $\gamma > 0$ and $\bm{\mu} = \bm{\Phi}^\top (\hat{\bm{\theta}}_{i,t} - \hat{\bm{\theta}}_{j,t})$ such as $\bm{\eta}$ and $\bm{\Phi}$ are the eigenvalues and the eigenvectors of the generalized eigenvalue problem:
    \begin{equation*}
        \bm{A}_{i,t} \bm{\Phi} = \bm{A}_{j,t} \bm{\Phi} \mathrm{diag}(\bm{\eta}) \quad \text{with} ~ \bm{\Phi}_i^\top \bm{A}_{j,t} \bm{\Phi}_i = 1 \enspace.
    \end{equation*}
\end{definition}

The parameter $\gamma$ in $\Psi$ is related to the ellipsoid radius and therefore depends on the level of confidence $\tilde{\delta}$ required for the separation.

\subsection{Separation time}

Two agents $i$ and $j$ may have a \emph{beneficial collaboration} as long as $\Psi(i, j, t) = 1$. Therefore, we consider collaboration sets of the form $\widehat{\mathcal{N}}_{i,t} = \{j ~|~ \Psi(i, j, t) = 1 \}$, and define the separation time between two agents as follows. Note that with our algorithm, once two agents are separated, they no longer share samples.

\begin{definition}[Separation Time]
\label{def:sep_time}

For two agents \( i \) and \( j \), the \emph{separation time} \( T_s(i, j) \) is defined as
\[
    T_s(i, j) = \min \left\{ t \in \mathbb{N} ~\middle|~ \Psi(i, j, t) = 0 \right\} \enspace,
\]

\end{definition}

The following result gives a lower bound on the separation time based on the parameter gap.

\begin{restatable}[Lower Bound on the Separation Time \( T_s \) Between Two Agents]{theorem}{thmb}
\label{th:lb_ts}

Let \( 0 < \delta < 1 \). Consider two agents \( i \) and \( j \), and suppose that \autoref{assum:bounded_norms} and \autoref{assum:sub_gaussian_noise} hold. Then, with probability at least \( 1 - \delta \), the separation time satisfies:
\[
    T_s(i, j) \geq \left\lceil \frac{8 \left( \mathrm{erf}^{-1}(1 - \delta) - \mathrm{erf}(\delta) \right)^2}{\| \bm{\theta}_i^* - \bm{\theta}_j^* \|_2} \right\rceil \enspace,
\]
where \( \mathrm{erf}(\cdot) \) denotes the Gauss error function.
\end{restatable}

\section{The Bandit Adaptive Sample Sharing algorithm}
\label{sec:bass_and_analysis}

\subsection{Description of the algorithm}
\label{subsec:bass}

Following~\autoref{def:ell_test}, we define a similarity graph $\mathcal{G}_t = (V, E_t)$, where $V = \{1, \ldots, N\}$ and $E_t = \{i, j ~|~ \Psi(i, j, t) = 1\}$, initialized as a complete graph. Therefore, agents $i$ and $j$ share their observations if and only if $(i, j) \in E_t$. The Bandit Adaptive Sample Sharing (BASS) Algorithm goes as follows.

\begin{figure}[t]
\scalebox{1.0}{
    \begin{minipage}{1.04\columnwidth }
    \begin{algorithm}[H]
        \caption{Bandit Adaptive Sample Sharing (\emph{BASS}) algorithm}
        \label{alg:bass}
        \textbf{Input}: \\
        The UCB parameter $\alpha$, the exploration parameter $\epsilon_e$ and the confidence level $\delta$. \\
        \textbf{Initialization}: \\
        For $i = 1, \ldots, N$ set $\bm{b}_{i,0} = \bm{0}_d$, $\bm{A}_{i,0} = \bm{I}_d$ \\ and the \emph{complete} graph $\mathcal{G}_0 = (V, E_0)$ ; \\[-1.em]
        \begin{algorithmic}[1]
            \FOR{$t=1,\dots, T$}
                    \FOR{$i \in V$}
                    \STATE Determine the neighbors $\widehat{\mathcal{N}}_{i, t-1}$ of agent $i$ ; \\ 
                    \STATE Update its current neighbor weights, set: \\            
                        \quad  $\ANithat = \bm{A}_{i,0} + \sum_{j \in \hat{\mathcal{N}}_{i, t-1}} (\bm{A}_{j,t-1} - \bm{A}_{j,0})$ , \\
                        \quad  $\bm{b}_{\Nithat} = \sum_{j \in \Nithat} \bm{b}_{j, t-1}$ , \\
                        \quad  $\thetaNithat = \ANithat^{-1} \bm{b}_{\Nithat}$ ; \\
                    \STATE Select $k_t \in \underset{k=1, \ldots, K}{\argmax} \quad \thetaNithat^\top \bm{x_k} + \beta_{\ANithat^{-1}}(\bm{x_k})$ \\
                    \quad with $\beta_{\ANithat^{-1}}(\bm{x}) = \alpha \sqrt{\bm{x}^\top \ANithat^{-1} \bm{x} \log(t)}$ \\
                    \STATE Pull $k_t$ and observe payoff $y_t$ ;
                    \STATE Update weights, set: \\
                        \quad $\bm{A}_{i, t} = \bm{A}_{i, t-1} + \bm{x}_{k_t}\bm{x}_{k_t}^\top$, \\
                        \quad $\bm{b}_{i, t} = \bm{b}_{i, t-1} + y_t\bm{x}_{k_t}$ ; \\
                \ENDFOR  
                \STATE Update graph $\mathcal{G}_t = (N, E_t)$:  \\
                    \quad Remove from $E_{t-1}$ all $(i, j)$ such as $\Psi(i, j, t) = 0$ ;
            \ENDFOR
        \end{algorithmic}
    \end{algorithm}
    \end{minipage}%
}
\end{figure}

 At each iteration $t$, for each agent $i \in V$\footnote{See the experiments in~\autoref{sec:exp} for the case of a single agent pulling an arm at each iteration.}, the shared estimator $\hat{\bm{\theta}}_{\mathcal{N}_i(t)}$ is updated and an arm is pulled according to OFUL policy with an exploration parameter $\alpha$. The agent then uses the observed rewards to update its local estimate $\hat{\bm{\theta}}_i$. Once every active agent has pulled an arm and updated its local estimate, the similarity graph $\mathcal{G}_t$ is updated according to~\autoref{def:ell_test}. The complete procedure is detailed in Algorithm~\ref{alg:bass}. 
 
 The overall complexity of the \emph{BASS} algorithm is $\mathcal{O}\left( T(N + Kd^2 + (N + 1)d^3) \right)$. We report the computational details and the associated complexity in the supplementary material~\autoref{sec:supp_implementation_details}.

Moreover, let \( C(T) \) denote the total communication cost up to time \( T \). In the worst-case scenario---where all \( N \) agents share identical bandit parameters and communicate their observations at each round---we have $C(T) \leq (N - 1) T$.

\subsection{Theoretical analysis}
\label{sec:theory}

\paragraph{Separation time upper-bound} \quad To facilitate this theoretical analysis, we consider a slightly modified version of the algorithm \emph{BASS} where the OFUL sampling is replaced with a uniform sampling with probability $\epsilon_e$, to ensure exploration in all directions. Empirically, we did not notice any difference with the above version.

\begin{restatable}[Upper Bound on the Separation Time \( T_s \)]{theorem}{thmc}
\label{th:ub_ts}

Let \( 0 < \delta < 1 \), and consider two agents \( i \) and \( j \). Let \( \epsilon_e > 0 \) satisfy
\[
    \epsilon_e \geq \frac{(2 + \sqrt{2})^2 R^2}{2 Q_{i,j}}, ~ \text{where} ~ Q_{i,j} = \frac{1}{K} \sum_{k=1}^{K} \left( \bm{x}_k^\top (\bm{\theta}_i^* - \bm{\theta}_j^*) \right)^2 \enspace.
\]
Then, under \autoref{assum:bounded_norms}, \autoref{assum:sub_gaussian_noise}, and following \autoref{alg:bass}, we have, with probability at least \( 1 - \delta \),
\[  
    T_s(i, j) \leq \left\lceil \frac{4 (2 + \sqrt{2})^2 R^2 \log \frac{1}{\delta}}{2 \epsilon_e Q_{i,j} - (2 + \sqrt{2})^2 R^2} \right\rceil \enspace.
\]
\end{restatable}

\paragraph{Cumulative pseudo-regret upper bound} \quad We focus on the particular case of collaboration between two agents, as it can be easily generalized to $N$ agents by considering all pairs. We focus on the cumulative regret at separation time, as the problem transitions to the well-known single agent setting afterward: $R_{i,T}^{\mathrm{collab}} = R_{i, 0, T_s(i,j)}^{\mathrm{collab}} + \sum_{t=T_s(i,j)}^T r_{i,t} \enspace$, with $R_{i, 0, T_s(i,j)}^{\mathrm{collab}} = \sum_{t=0}^{T_s(i,j)} r_{i,t}$.

\begin{restatable}[Individual Regret During the Collaboration Phase]{theorem}{thmd}
\label{th:upper_bound_general}

Let \( T_s = T_s(i, j) \) denote the separation time introduced in \autoref{def:sep_time}. Under \autoref{assum:bounded_norms} and \autoref{assum:sub_gaussian_noise}, the cumulative regret of agent \( i \) during the collaboration phase is bounded as:
\[
    R_{i, 0, T_s}^{\mathrm{collab}} \leq \mu(\delta, d, \gamma) \cdot \nu(\delta, d, T_s) + \frac{T_s}{2} \left( \bm{\theta}_i^* - \bm{\theta}_j^* \right)^\top \bm{x}_i^* \enspace,
\]
where
\[
    \mu(\delta, d, \gamma) = \frac{1}{2} + \frac{\gamma}{4} + \frac{1}{2\sqrt{2}} \sqrt{1 + \frac{d \log 2}{2 \log \frac{1}{\delta}}} \enspace,
\]
and
\[
    \nu(\delta, d, T_s) = \sqrt{4 \beta(\delta, \bm{A}_{T_s})^2 T_s d \log \left(1 + \frac{T_s}{d} \right)} \enspace.
\]
\end{restatable}

With $\nu(\delta, d, T_s)$ we recognize the upper bound in the case of a single agent as derived in~\citet{Abbasi2011}, thus $\mu(\delta, d, \gamma)$ represents the collaborative gain and $\frac{T_s}{2} (\bm{\theta}_i^* - \bm{\theta}_j^*)^\top \bm{x}_i^*$ the result of the induced bias. Setting $\gamma < 2 - \sqrt{2}\sqrt{1 + (d \log 2) / (2 \log \frac{1}{\delta})}$, we have $\mu(\delta, d, \gamma) < 1$.

\paragraph{Network cumulative pseudo-regret upper bound} \quad Looking at the cumulative regret of all agents, we can further refine our upper bound. Let us set $\bar{R}_{0, T_s}^{\mathrm{collab}} = \frac{1}{2} \sum_{i=1}^{2} R_{i, 0, T_s}^{\mathrm{collab}}$, the averaged cumulative pseudo-regret.

\begin{restatable}[Regret During the Collaboration Phase]{theorem}{thme}
\label{th:upper_bound_averaged_general}

Let \( T_s = T_s(i, j) \) denote the separation time introduced in \autoref{def:sep_time}. Under \autoref{assum:bounded_norms} and \autoref{assum:sub_gaussian_noise}, the cumulative regret of all agents during the collaboration phase is bounded as:
\[
    \bar{R}_{0, T_s}^{\mathrm{collab}} \leq \mu'(\delta, d) \cdot \nu(\delta, d, T_s) + \frac{T_s}{4} \left( \bm{\theta}_i^* - \bm{\theta}_j^* \right)^\top \left( \bm{x}_i^* - \bm{x}_j^* \right) \enspace,
\]
where
\[
    \mu'(\delta, d) = \frac{1}{\sqrt{2}} \sqrt{1 + \frac{d \log 2}{2 \log \frac{1}{\delta}}} \enspace,
\]
and \( \nu(\delta, d, T_s) \) is as previously defined.
\end{restatable}

Similarly to~\autoref{th:upper_bound_general} we recognize the upper bound in the case of a single agent, thus $\mu'(\delta, d)$ represents the collaborative gain and $\frac{T_s}{4} (\bm{\theta}_i^* - \bm{\theta}_j^*)^\top (\bm{x}_i^* - \bm{x}_j^*)$ bias-related quantity.

\paragraph{Regret analysis with clustered parameters} \quad Most existing algorithms rely on the assumption that the linear bandit parameters are clustered around a few centroids. We thus extend our analysis to this specific assumption, which is formalized as:
\begin{assumption}[Clustering Structure]
\label{assum:clustering_struct}
We assume that there exist \( M \ll N \) distinct bandit parameters \( \{ \bm{\theta}_1^*, \ldots, \bm{\theta}_M^* \} \subset \mathbb{R}^d \) such that, for all agents \( i \in \{1, \ldots, N\} \), the true parameter satisfies:
\[
    \bm{\theta}_i^* \in \{ \bm{\theta}_1^*, \ldots, \bm{\theta}_M^* \} \enspace.
\]
\end{assumption}

Under the assumption of clustered parameters, the collaboration may \emph{always be beneficial}. We also introduce additional notation: for a given agent $i$, we define its true neighborhood (\ie the set of agents within its cluster) as $\mathcal{N}_i$. We also introduce the cardinalities $N_i = \rho_{m(i)} N = |\mathcal{N}_i|$ and $\hat{N}_{i, t} = |\widehat{\mathcal{N}_{i,t}}|$. Finally, we denote by $N_{i, t}^e$ the gap between both cardinalities, that is, $N_{i, t}^e = |\hat{N}_{i, t} -  N_{i, t}|$. The following result upper bounds the number of misassigned agents.

\begin{restatable}[Expected Number of Misassigned Agents]{lemma}{thmf}
\label{th:neighboor_error}

Let \( 0 < \delta < 1 \). Under \autoref{assum:bounded_norms}, \autoref{assum:sub_gaussian_noise}, and \autoref{assum:clustering_struct}, we have, with probability at least \( 1 - \delta \):
\[
    \mathbb{E} \left[ N_i^e(t) \right] \leq n_i^e(t) \enspace,
\]
where
\[
    n_i^e(t) = \left\lceil \frac{N \sqrt{\det(\bm{A}_t)}}{ \exp\left( \frac{1}{2} \left( \Delta^{\min}_{\bm{A}_t} - \gamma \beta(\delta, \bm{A}_t) \right)^2 \right) } - \rho_{m(i)} N \right\rceil \enspace,
\]
and
\[
    \Delta^{\min}_{\bm{A}_t} = \min_{i \neq j} \| \bm{\theta}_i^* - \bm{\theta}_j^* \|_{\bm{A}_t} \enspace.
\]
\end{restatable}

With this clustering error bound, we can derive an upper bound on the cumulative pseudo-regret denoted $R_{i, 0, T}^{\mathrm{cluster}}$. We provide additional theoretical results in the supplementary material~\autoref{sec:supp_additional_analysis}.

\begin{restatable}[Cumulative Pseudo-Regret with Clustered Agents]{theorem}{thmg}
\label{th:upper_bound_clustering}

Under \autoref{assum:bounded_norms}, \autoref{assum:sub_gaussian_noise}, and \autoref{assum:clustering_struct}, after \( T \) iterations of the \emph{BASS} algorithm, the cumulative pseudo-regret of agent \( i \) satisfies:
\[
    R_{i, 0, T}^{\mathrm{cluster}} \leq \mu''(\delta, d, N) \cdot \nu(\delta, d, T) + \frac{4L}{\delta \rho_{\min} N} \cdot C_i(T_c) \enspace,
\]
where:
\[
    C_i(T_c) = \sum_{t=1}^{\min(T, T_c)} n_i^e(t) \enspace,
\]
\[
    \mu''(\delta, d, N) = \frac{1}{\sqrt{ \rho_{\min} N}} \sqrt{1 + \frac{d \log N}{2 \log \frac{1}{\delta}}} \enspace,
\]
\[
    T_c = \max_j T_s(i, j), \quad \rho_{\min} = \min_j \rho_{m(j)} \enspace,
\]
and \( \nu(\delta, d, T) \) is as previously defined.
\end{restatable}
\paragraph{Discussion} \quad To better interpret this result, we derive a lower bound on the cumulative pseudo-regret in the idealized case where all agents share the same bandit parameter and form a single, known cluster, we obtain:
\[
    R_{i, 0, T}^{\mathrm{cluster}} \geq \frac{\nu(\delta, d, T)}{N} \enspace.
\]
This bound emphasizes the additional cost term \( \frac{4L}{\delta \rho_{\min} N} \cdot C_i(T_c) \) in~\autoref{th:upper_bound_clustering}, which accounts for the effort required to correctly identify agent clusters.

Additionally, we summarize in~\autoref{tab:bounds} the upper bounds available for the concurrent algorithms considered under the structure assumption of the clustered agents. We provide the derivation steps for this table and include an additional oracle case in the supplementary materials~\autoref{sec:supp_additional_analysis}.

\begin{table}[ht]
    \centering
    \scalebox{0.825}{
        \begin{tabular}{ c | c | c }
           & \thead{Cumululative \\ pseudo regret} & \thead{Expected number \\ of misassigned agents} \\
          \hline
          DynUCB &  not available  & not available  \\
          \hline
          CLUB & $\mathcal{O}\left( d \sqrt{\frac{M}{N}} \sqrt{T} \log(T)\right)$ & not available  \\
          \hline
          SCLUB & $\mathcal{O}\left( d \sqrt{\frac{M}{N}} \sqrt{T} \log(T) \right)$  & not available  \\
          \hline
          CMLB & $\mathcal{O}\left( \frac{\sqrt{d}}{\sqrt{\rho^{\min} N}} \sqrt{T} \log(T)^2 \right)$ & not available  \\
          \hline
          \textbf{BASS} & $\bm{\mathcal{O}\left( \frac{\sqrt{d}}{\sqrt{\rho^{\min} N}}  \sqrt{T} \log(T / d)\right)}$ & $\bm{\mathcal{O}\left( \frac{N}{\sqrt{T}^{~\gamma^2 R^2 - 1}} \right)}$ \\
        \end{tabular}
    }
    \caption{Summary of available upper-bounds for clustered agents.}
    \label{tab:bounds}
\end{table}

Our analysis of the cumulative pseudo-regret with clustered agents establishes the tightest upper bounds, incorporating a $\sqrt{d} / \sqrt{\rho^{\min} N}$ constant term and a $T / d$ term within the logarithm. Furthermore, to our knowledge, this is the first work to provide an analysis of the clustering error of the algorithm.

\section{Experiment}
\label{sec:exp}

All upper bounds include an \emph{gain} term compared to the single agent case and a \emph{loss} term from the introduced bias. In this section, we demonstrate that the combination of these two terms results in an overall \emph{improved regret minimization} on both synthetic and real data experiments. The experiments were run in Python on $50$ '\emph{Intel Xeon @ 3.20 GHz}' CPUs and lasted a week. The code is publicly available and can be found at \href{https://github.com/hcherkaoui/collaborative_bandits}{this repository}.

\begin{figure}[ht]
    \centering
        \begin{tabular}{lr} 
            \includegraphics[width=0.73\linewidth,trim= 0 0 0 0,clip]{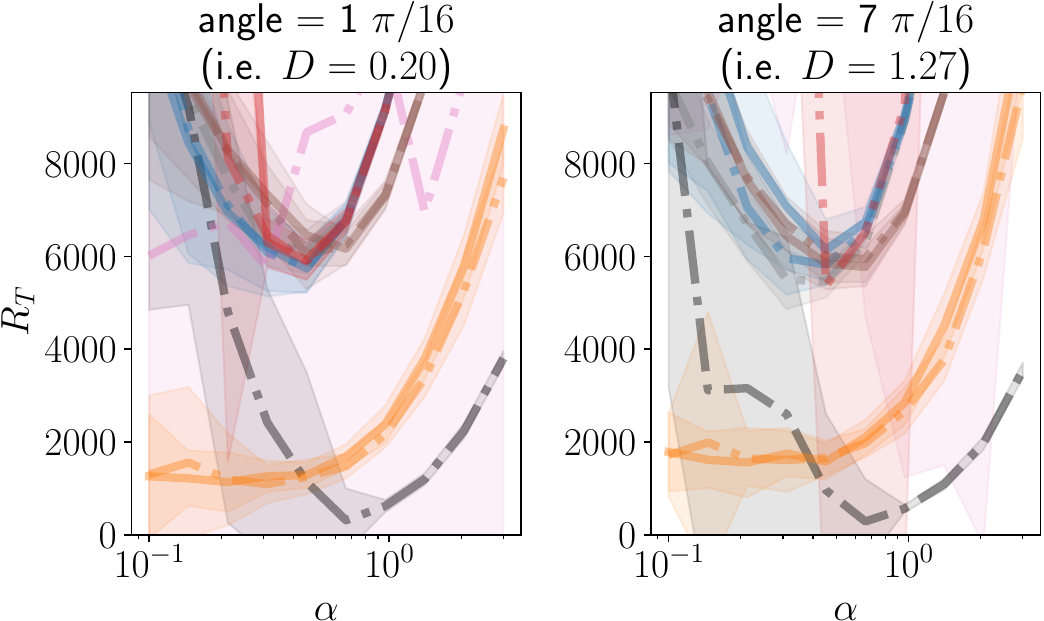} &
            \includegraphics[width=0.24\linewidth,trim= 2 2 2 2,clip]{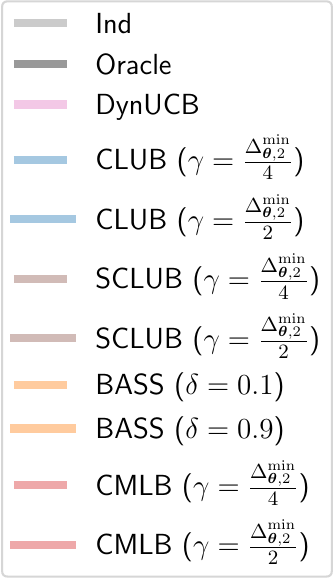} \\
        \end{tabular}
    \caption{Comparison of the averaged evolution of the cumulative regret last value $R_T$ \wrt{} the UCB parameter $\alpha$ for the different synthetic environments considered.}%
    \label{fig:figure_regret_evolution_Rt_synth_data}
\end{figure}

\paragraph{Benchmark on synthetic data} \quad To match the asynchronous pulling commonly used in the literature, we consider that only one agent, randomly chosen by the environment, pulls an arm at each iteration. Similarly, to match the common assumption of the clustering structure, we consider~\autoref{assum:clustering_struct} for the experiments. We propose to benchmark the algorithm on two variations of a synthetic environment. We consider $M=3$ clusters of the same size with bandit parameter $(\bm{\theta}_m)_{m \in \{1, \ldots, M\}}$ defined as: $\forall 1 \leq q \leq \lceil \frac{M}{2} ~ \rceil \bm{\theta}_{2q-1} = \bm{e}_q$, with $(\bm{e}_i)_i$ being the canonical basis and $\bm{\theta}_{2q}$ having its $q$-th entry being $\cos{\omega}$, its $(q+1)$-th entry being $\sin{\omega}$ and all the other entries being $0$, with $\omega \in \{\pi/16, 7\pi/16\}$. The angle $\omega$ induces a problem complexity $\Delta^{\min}_{2}$, denoted $D$ in~\autoref{fig:figure_regret_evolution_Rt_synth_data}, of $0.2$ in the first scenario and $1.27$ in the second.

We consider $K = 5 \times M$ arms of dimension $d=10$. The algorithms are set to iterate for $T=50000$. We set the number of agents to $N=100$ and corrupt the reward observation with a centered and normalized Gaussian noise. As baselines, we choose \emph{Ind}, which proposes to run the agents independently and \emph{Oracle} which knows the true group structure and shares observations within the same cluster. We also compare our performance to state-of-the-art algorithms: \emph{DynUCB}, \emph{CLUB}, \emph{SCLUB}, \emph{CMLB}. For each algorithm, we consider two sets of hyper-parameters for which the result is shown in the `dashed' and `solid' lines. For each method, we line-search the $\alpha$ parameter. We set our hyperparameter $\gamma$ to $2$ to match~\citet{Gilitschenski2012}. The experiment is run $50$ times to average across runs. We detail the experimental setting and a description of the concurrent algorithms in~\autoref{sec:supp_experiment_details}.

We report an additional case with $M=6$ and an additional synthetic benchmark with Gaussian arms and bandit parameters in~\autoref{sec:supp_additional_experiments}.

\begin{table}
    \centering
    \scalebox{0.8}{%
\hspace*{-0.55cm} 
\begin{tabular}{l|ccc} 
   \textbf{Cumul. regret} & $\Delta^{\min}_{\bm{\theta}, 2}=0.2$ & $\Delta^{\min}_{\bm{\theta}, 2}=1.27$ \\ 
    \midrule 
    Ind & 6100.7 +/- 468.4 & 5465.0 +/- 352.7 \\ 
    \emph{Oracle} & 318.5 +/- 684.0 & 292.9 +/- 914.3 \\ 
    DynUCB & 5765.5 +/- 10439.5 & 15081.6 +/- 15299.7 \\ 
    CLUB ($\gamma=\Delta^{\min}_{\bm{\theta}, 2}/4$) & 5772.5 +/- 532.4 & 5822.7 +/- 430.6 \\ 
    CLUB ($\gamma=\Delta^{\min}_{\bm{\theta}, 2}/2$) & 5739.4 +/- 521.5 & 6138.1 +/- 659.8 \\ 
    SCLUB ($\gamma=\Delta^{\min}_{\bm{\theta}, 2}/4$) & 6038.5 +/- 744.8 & 5949.3 +/- 501.0 \\ 
    SCLUB ($\gamma=\Delta^{\min}_{\bm{\theta}, 2}/2$) & 6165.6 +/- 366.0 & 5775.3 +/- 416.8 \\ 
    BASS ($\delta=0.1$) & \textbf{1096.8 +/- 411.9} & 1617.5 +/- 689.8 \\ 
    BASS ($\delta=0.9$) & 1140.5 +/- 641.4 & \textbf{1563.2 +/- 750.7} \\ 
    CMLB ($\gamma=\Delta^{\min}_{\bm{\theta}, 2}/4$) & 5895.3 +/- 382.5 & 5372.1 +/- 7575.0 \\ 
    CMLB ($\gamma=\Delta^{\min}_{\bm{\theta}, 2}/2$) & 5918.1 +/- 431.3 & 20948.5 +/- 5355.5 
\end{tabular} 
} 

    \caption{Comparison of the averaged cumulative regret last value $R_T$ for the different synthetic environments.}
    \label{tab:table_R_T_1_benchmark_synth__M_3} 
\end{table}

\begin{figure}[ht]
    \centering
        \includegraphics[width=0.72\linewidth,trim= 0 165 0 0,clip]{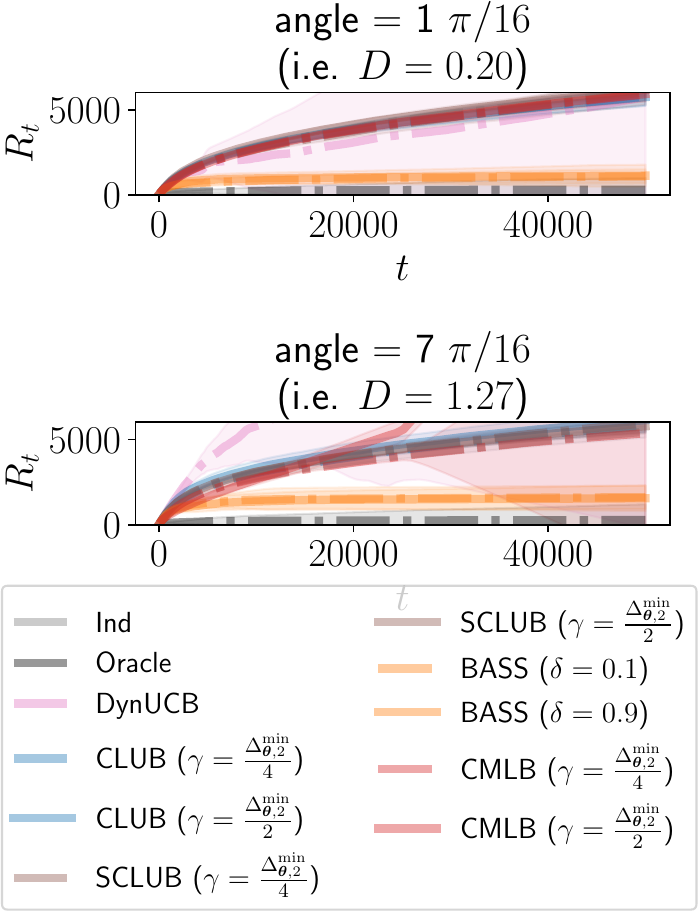}
    \caption{Comparison of the averaged evolution of the cumulative regret last value $R_t$ for the different synthetic environments considered.}%
    \label{fig:figure_regret_evolution_Rt_synth_data__n_thetas_3}
\end{figure}

In~\autoref{tab:table_R_T_1_benchmark_synth__M_3} (\resp{}~\autoref{fig:figure_regret_evolution_Rt_synth_data__n_thetas_3}), we report the cumulative regret last value $R_T$ (\resp{} $R_t$) for all concurrent and baseline methods. First, we note the expected behavior of the baselines: \emph{Oracle} provides the best performance for all scenarios, and the \emph{Ind} algorithm provides intermediate performance, as the agent does not share its observations, but avoids introducing a detrimental estimation bias. We find that our approach, \emph{BASS}, significantly outperforms all the other algorithms. In particular, we notice that our approach always results in a gain in comparison to \emph{Ind}, as underlined by~\autoref{sec:bass_and_analysis}. Moreover, in~\autoref{fig:figure_regret_evolution_Rt_synth_data} we show the cumulative regret last value, $R_T$, of the evolution \wrt{} the UCB parameter $\alpha$. We can see that our approach systematically provides a lower regret than the state-of-the-art algorithms, underscoring the robustness of \emph{BASS}.

\begin{table}
    \centering
    \scalebox{0.8}{%
\hspace*{-0.57cm} 
\begin{tabular}{l|ccc} 
   \textbf{Clustering score} & $\Delta^{\min}_{\bm{\theta}, 2}=0.2$ & $\Delta^{\min}_{\bm{\theta}, 2}=1.27$ \\ 
    \midrule 
    Ind & 0.0 +/- 0.0 & 0.0 +/- 0.0 \\ 
    \emph{Oracle} & 1.0 +/- 0.0 & 1.0 +/- 0.0 \\ 
    DynUCB & \textbf{0.7 +/- 0.1} & \textbf{0.9 +/- 0.2} \\ 
    CLUB ($\gamma=\Delta^{\min}_{\bm{\theta}, 2}/4$) & 0.0 +/- 0.0 & 0.0 +/- 0.0 \\ 
    CLUB ($\gamma=\Delta^{\min}_{\bm{\theta}, 2}/2$) & 0.0 +/- 0.0 & 0.0 +/- 0.0 \\ 
    SCLUB ($\gamma=\Delta^{\min}_{\bm{\theta}, 2}/4$) & 0.0 +/- 0.0 & 0.0 +/- 0.0 \\ 
    SCLUB ($\gamma=\Delta^{\min}_{\bm{\theta}, 2}/2$) & 0.0 +/- 0.0 & 0.0 +/- 0.0 \\ 
    BASS ($\delta=0.1$) & \textbf{0.7 +/- 0.0} & \textbf{0.9 +/- 0.0} \\ 
    BASS ($\delta=0.9$) & \textbf{0.7 +/- 0.0} & 0.8 +/- 0.0 \\ 
    CMLB ($\gamma=\Delta^{\min}_{\bm{\theta}, 2}/4$) & 0.0 +/- 0.0 & 0.0 +/- 0.2 \\ 
    CMLB ($\gamma=\Delta^{\min}_{\bm{\theta}, 2}/2$) & 0.0 +/- 0.0 & 0.6 +/- 0.1
\end{tabular} 
} 

    \caption{Comparison of the averaged F$1$ score of the estimated graph $\mathcal{G}_T$ for the different synthetic environments.}
    \label{tab:table_clustering_score_1_benchmark_synth__M_3} 
\end{table}

In~\autoref{tab:table_clustering_score_1_benchmark_synth__M_3}, we show the final clustering score value for all clustering algorithms. To quantify the degree of agreement between the estimated clusters/neighborhoods and the ground truth, we compute the $\mathrm{F}_1$-score on the graph $\mathcal{G}_t$ w.r.t.\ the true cluster graph $\mathcal{G}$. Details of the score computation are reported in~\autoref{sec:supp_experiment_details}. We observe that our approach always achieves the best performances. Moreover, most of the concurrent methods recover the clustering structure poorly. Since \citet{Nguyen2014, Gentile2014, Li2019, Ghosh2022} do not provide any experiment to verify this aspect, these results are not surprising.

\paragraph{Benchmark on real data}\label{subsec:bench_real_data} \quad To complete the performance study of our method, we propose a second benchmark with real datasets. We consider two public datasets of ranking scenarios: \emph{MovieLens} and \emph{Yahoo!} dataset. Since \emph{Oracle} is not available for real data, we consider all other previous algorithms except it, keeping the same experimental setting. Details of the experimental setting are reported in~\autoref{sec:supp_experiment_details}.

\begin{table}
    \centering
    \scalebox{0.8}{%
\hspace*{-0.55cm} 
\begin{tabular}{l|ccc} 
   \textbf{Cumul. regret} & Movie Lens & Yahoo \\ 
    \midrule 
    Ind & 11825.2 +/- 594.1 & 14800.5 +/- 876.0 \\ 
    DynUCB ($M=5$) & 16835.0 +/- 5429.9 & 14689.5 +/- 1912.7 \\ 
    DynUCB ($M=20$) & 17256.0 +/- 2314.2 & 16860.5 +/- 1625.3 \\ 
    CLUB ($\gamma=0.1$) & 11504.2 +/- 509.3 & 13321.5 +/- 4228.8 \\ 
    CLUB ($\gamma=1.0$) & 11804.8 +/- 562.4 & 13536.5 +/- 6548.2 \\ 
    SCLUB ($\gamma=0.1$) & 11425.2 +/- 764.0 & 13663.0 +/- 1444.0 \\ 
    SCLUB ($\gamma=1.0$) & 11085.5 +/- 701.3 & 13751.5 +/- 820.1 \\ 
    BASS ($\delta=0.1$) & \textbf{10273.2 +/- 686.3} & \textbf{12456.0 +/- 2096.1} \\ 
    BASS ($\delta=0.9$) & 10840.2 +/- 831.7 & 12533.5 +/- 1959.5 \\ 
    CMLB ($\gamma=0.1$) & 14944.2 +/- 341.6 & 16812.5 +/- 405.1 \\ 
    CMLB ($\gamma=1.0$) & 12010.8 +/- 669.4 & 16757.5 +/- 607.2
\end{tabular} 
} 

    \caption{Comparison of the averaged cumulative regret last value $R_T$ for the two datasets.}
    \label{tab:table_R_T_2_benchmark_real} 
\end{table}

In \autoref{tab:table_R_T_2_benchmark_real}, we show the final cumulative regret $R_T$ for the considered datasets. We observe that our approach again outperforms all the others in both scenarios. The overall performance comparison between the algorithms remains consistent with the synthetic experiment and confirms the good behavior of our algorithm; additional results are reported in~\autoref{sec:supp_additional_experiments}.

\section{Conclusion}
\label{sec:ccl}

This paper explores collaboration as a sample sharing process, highlighting the benefits of introducing a bias, provided that it remains smaller than the current uncertainty. We derive a collaboration condition and propose an algorithm that takes advantage of sample sharing. Our approach has been extensively analyzed both theoretically and empirically, highlighting the key quantities involved in the process and demonstrating strong empirical performance on both synthetic and real-world datasets.

To our knowledge, this is the first work to have comprehensively analyzed the introduction of bias in this context or established an efficient condition and collaboration scheme with anisotropic testing. In line with these research directions, designing an arm-pulling strategy aimed at efficiently estimating the similarities could be a promising avenue.

\clearpage
\newpage

\section*{Impact Statement}
\label{sec:impact}

This paper presents work whose goal is to advance the field of Machine Learning. There are many potential societal consequences of our work, none which we feel must be specifically highlighted here.

\bibliographystyle{icml2025}
\bibliography{bibliography}

\clearpage
\newpage
\appendix
\onecolumn 

\section{Appendix: table of contents}
\label{sec:supp_table_of_contents}

\begin{enumerate}[align=left,itemsep=0.5cm,label=\Alph*]
    \item Appendix: Notations summary \dotfill \pageref{sec:supp_notations}  
    \item Appendix: A simple two agents problem: Details on~\autoref{subsec:main_idea} \dotfill \pageref{sec:supp_pre_analysis_theta_est}  
    \item Appendix: Technical lemmas \dotfill \pageref{sec:supp_technical_lemmas}  
    \item Appendix: Bandit parameter estimation error bounding: Details on~\autoref{th:RHS_est_theta} \dotfill \pageref{sec:supp_single_agent_regret_minimization}  
    \item Appendix: Instantaneous regret upper bounds: Proof of~\autoref{lem:regret_bound_inst} \dotfill \pageref{sec:supp_instantaneous_regret_ub}  
    \item Appendix: Collaboration condition: Details on Equation~\autoref{eq:stop_crit} \dotfill \pageref{sec:supp_initial_sep_criteria}  
    \item Appendix: Ellipsoid separation under synchronous pulling: Proof of~\autoref{cor:ell_test} \dotfill \pageref{sec:supp_gili_equivalence}  
    \item Appendix: Ellipsoid separation test function: Details on~\autoref{def:ell_test} \dotfill \pageref{sec:supp_our_test_equivalence}  
    \item Appendix: Lower bound on the separation time: Proof of~\autoref{th:lb_ts} \dotfill \pageref{sec:supp_Ts_lb}  
    \item Appendix: Upper bound on the separation time: Proof of~\autoref{th:ub_ts} \dotfill \pageref{sec:supp_Ts_ub}  
    \item Appendix: Cumulative pseudo-regret upper bound: Proof of~\autoref{th:upper_bound_general} \dotfill \pageref{sec:supp_first_cumul_regret_ub}  
    \item Appendix: Network cumulative pseudo-regret upper bound: Proof of~\autoref{th:upper_bound_averaged_general} \dotfill \pageref{sec:supp_second_cumul_regret_ub}  
    \item Appendix: Expected number of misassigned agents: Proof of~\autoref{th:neighboor_error} \dotfill \pageref{sec:supp_clustering_error}  
    \item Appendix: Cumulative pseudo regret with clustered agents: Proof of~\autoref{th:upper_bound_clustering} \dotfill \pageref{sec:supp_third_cumul_regret_ub}  
    \item Appendix: Additional theoretical analysis: \dotfill \pageref{sec:supp_additional_analysis}  
    \item Appendix: Implementation details \dotfill \pageref{sec:supp_implementation_details}  
    \item Appendix: Experimental details \dotfill \pageref{sec:supp_experiment_details}  
    \item Appendix: Additional details \dotfill \pageref{sec:supp_additional_experiments}  
\end{enumerate}

\clearpage
\newpage

\section{Appendix: Notations summary} 
\label{sec:supp_notations}

\renewcommand{\arraystretch}{1.1}

\scalebox{0.95}{
    \begin{tabularx}{\textwidth}{X|X}
        \textbf{Variables and parameters} & \textbf{Details} \\
        \hline
         $d \in \mathbf{N}$ & Dimension \\
        \hline
         $K \in \mathbf{N}$ & Number of arms \\
        \hline
         $N \in \mathbf{N}$ & Number of agents \\
        \hline
         $M \in \mathbf{N}$ & True number of clusters \\
        \hline
         $\rho_i N \in \mathbf{N}$ & Number of agents within the neighborhood of agent $i$ \\
        \hline
         $T \in \mathbf{N}$ & Total number of iterations \\
        \hline
         $T_s(i, j) \in \mathbf{N}$ & Separation iteration of agent $i$ and $j$ \\
        \hline
         $T_c = \max_j T_s(i, j)$ & Iteration at which the cluster is adequately estimated  \\
        \hline
         $\mathcal{X} = \{\forall k \in [K] ~|~ \bm{x}_k \in \mathbf{R}^d \text{ with } \|\bm{x}\|_2 \leq 1 \}$ & Arm set \\
        \hline
         $\bm{\theta}_i^* \in \mathbf{R}^d$ & True bandit parameter of agent $i$ \\
        \hline
         $L \in \mathbf{R}^{*+}$ & Bound of the bandit parameter norm \ie{} $\forall i ~ \|\bm{\theta}_i^*\|_2 \leq L$ \\
        \hline
         $\eta \in \mathbf{R}$ & Reward observation noise \\
        \hline
         $R \in \mathbf{R}^{*+}$ & Sub-Gaussianity parameter of the noise \\
        \hline
         $y_t= \bm{x}_t^\top \bm{\theta}^* + \eta \in \mathbf{R}$ & Reward signal at iteration t \\
        \hline
         $\mathcal{F}_t$ & $\sigma$-algebra generated by $(\bm{x}_1, y_1, \ldots, \bm{x}_{t-1}, y_{t-1}, \bm{x}_{t})$ \\
        \hline
         $r_t = \max_k \bm{\theta}^{* \top} \bm{x}_{k} - \bm{\theta}^{* \top} \bm{x}_{t} \in \mathbf{R}^{+}$ & Instantaneous regret \\
        \hline
         $\bm{A}_{i, t} =  \sum_{s = 1}^t \bm{x}_{i,s} \bm{x}_{i,s}^\top \in \mathbf{R}^{d \times d}$ & Design matrix at iteration $t$ of agent $i$ \\
        \hline
         $\ANithat = \bm{A}_{i,0} + \sum_{j \in \hat{\mathcal{N}}_{i, t-1}} (\bm{A}_{j,t-1} - \bm{A}_{j,0}) \in \mathbf{R}^{d \times d}$ & Design matrix at iteration $t$ of neighborhood $i$ \\
        \hline
         $\bm{b}_{i, t} = \sum_{s=1}^t + y_{i,t}\bm{x}_{i,t} \in \mathbf{R}^d$ & Regressand for the estimation of the bandit parameter of the agent $i$ at iteration $t$ \\
        \hline
         $\bm{b}_{\Nithat} = \sum_{j \in \Nithat} \bm{b}_{j, t-1} \in \mathbf{R}^d$ &  Regressand for the estimation of the bandit parameter of the neighborhood $i$ at iteration $t$ \\
        \hline
         $\thetaNithat = \ANithat^{-1} \bm{b}_{\Nithat} \in \mathbf{R}^d$ & Bandit parameter estimation from neighborhood $i$ \\
        \hline
         $\bm{\hat{\theta}}_i \in \mathbf{R}^d$ & Bandit parameter estimation from agent $i$ \\
        \hline
        $\delta$ & Confidence level parameter \\
        \hline
         $\mathcal{C}_\delta(\hat{\bm{\theta}}_{t})$ & Confidence region of the estimated bandit parameter \\
        \hline
         $\beta(\delta, t) = R \sqrt{2 \log{\frac{1}{\delta}} +  \log{\text{det}(\bm{A}_{t})}} \in \mathbf{R}^{*+}$ & Confidence level of set of the previous region \\
        \hline
         $\widehat{\mathcal{N}}_i(t) = \{j ~|~\Psi(i,j) = 1\}$ & Neighborhood of agent $i$ \\
        \hline
         $N_i^e(t)$ & Number of agents wrongly assigned to neighborhood $i$ at iteration $t$ \\
        \hline
         $\Delta^{i, j}_{2} = \|\bm{\theta}_i^* - \bm{\theta}_j^*\|_2 \in \mathbf{R}^{*+}$ & Bandit parameters Euclidean distance \\
        \hline
         $\Delta^{\min}_{2} = \underset{i, j ~ i \neq j}{\min} \|\bm{\theta}_i^* - \bm{\theta}_j^*\|_2 \in \mathbf{R}^{*+}$ & Isotrope complexity of the problem \\
        \hline
         $\Delta^{i, j}_{\bm{M}} = \| \bm{\theta}^*_i - \bm{\theta}^*_j \|_{\bm{M}} \in \mathbf{R}^{*+}$ & Bandit parameters Mahalanobis distance weighted by matrix $\bm{M} \in \mathbf{R}^{d \times d}$ \\
        \hline
         $\Delta^{\min}_{\bm{M}} = \underset{i, j \quad i \neq j}{\min} \|\bm{\theta}_i^* - \bm{\theta}_j^*\|_{\bm{M}} \in \mathbf{R}^{*+}$ & Anisotrope complexity of the problem weighted by matrix $\bm{M} \in \mathbf{R}^{d \times d}$ \\
        \hline
         $E_t = \{(i, j) ~|~ \Psi(i,j) = 0 \}$ & Edges modelling the interaction between the agents \\
        \hline
         $V = \{1 \ldots N \}$ & Set of all agents \\
        \hline
         $G_t = (V, E_t)$ & Estimated agent graph at iteration $t$ \\
        \hline
         $\mathcal{I}_t \subset V$ & Agent pulling set at iteration $t$ \\
        \hline
         case $\mathcal{I}_t  = \{i\}$ & Asynchronous pulling (with $i$ draw randomly from $V$) \\
        \hline
         case $\mathcal{I}_t = V$ & Synchronous pulling
    \label{tab:notations}
    \end{tabularx}
}

\clearpage
\newpage

\section{Appendix: A simple two agents problem} 
\label{sec:supp_pre_analysis_theta_est}

\subsection{Preliminary theoretical results}

We detail here the bounding of estimation error of $\bm{\theta}_1^{*}$. From the series of noisy projections $(y_{1,s})_{s=1}^t$, with $y_{1,s} = {\bm{\theta}_1^{*}}^\top \bm{x}_{1,s} + \eta_{1,s}$ with $(\bm{x}_{1,s})_{s=1}^t \in \mathcal{X}^t$ a fixed series of arms. We consider the observation matrix to be invertible; the proof remains similar to the case involving the Moore-Penrose inverse. We can cancel the gradient of the OLS estimator to obtain:
\begin{align*}
   \hat{\bm{\theta}}_t &= \bm{\theta}_1^* + \bm{A}_t^{-1} \sum_{s=1}^t \bm{x}_{1, s} \eta_{1, s} \quad \text{with} \quad \bm{A}_t = \sum_{s=1}^t \bm{x}_{1, s} \bm{x}_{1, s}\top \\
   \left\| \hat{\bm{\theta}}_t - \bm{\theta}_1^* \right\|_{\bm{A}_t} &= \left\| \sum_{s=1}^t \bm{x}_{1, s} \eta_{1, s} \right\|_{\bm{A}_t^{-1}} \\
   \left\| \hat{\bm{\theta}}_t - \bm{\theta}_1^* \right\|_{\bm{A}_t} &\leq \beta(\delta, \bm{A}_t) \quad \text{as defined in~\autoref{th:RHS_est_theta}}
\end{align*}

Similarly, for the collaboration case we have:
\begin{align*}
    \hat{\bm{\theta}}_t^{\mathrm{collab}} &= \frac{1}{2}(\bm{\theta}_1^* + \bm{\theta}_2^*) + \bm{A}_t^{-1} \sum_{s=1}^t \bm{x}_{1, s} (\eta_{1, s} + \eta_{2, s}) \\
    \left\| \hat{\bm{\theta}}_t^{\mathrm{collab}} - \bm{\theta}_1^* \right\|_{\bm{A}_t} &= \left\| \frac{1}{2}(\bm{\theta}_2^* - \bm{\theta}_1^*) + \frac{1}{2} \bm{A}_t^{-1} \sum_{s=1}^t \bm{x}_{1, s} (\eta_{1, s} + \eta_{2, s}) \right\|_{\bm{A}_t} \\
    \left\| \hat{\bm{\theta}}_t^{\mathrm{collab}} - \bm{\theta}_1^* \right\|_{\bm{A}_t} &\leq \frac{1}{2} \left\| \bm{\theta}_2^* - \bm{\theta}_1^* \right\|_{\bm{A}_t} + \frac{1}{\sqrt{2}} \beta(\delta, \bm{A}_t)  \quad \text{as defined in~\autoref{th:RHS_est_theta}}
\end{align*}

Using the two previous error bounds, we compute the ratio of the bounding terms and compare it to $1$ to gain insight into the collaboration condition, leading to:
\begin{align*}
    \frac{\frac{1}{2} \left\| \bm{\theta}_j^* - \bm{\theta}_i^* \right\|_{\bm{A}_t} + \frac{1}{\sqrt{2}} \beta(\delta, \bm{A}_t)}{\beta(\delta, \bm{A}_t)} &\leq 1 \\
    \left\| \bm{\theta}_1^{*} -  \bm{\theta}_2^{*} \right\|_{\bm{A}_t} &\leq (2 - \sqrt{2}) \beta(\delta, \bm{A}_t)
\end{align*}

\subsection{Empirical illustration}

To generate~\autoref{fig:illustration}, we consider a simple case with $d = 2$ and a set of two arms $\mathcal{X} = \{\bm{e}_1, \bm{e}_2\}$, where $\bm{e}_1$ and $\bm{e}_2$ are the canonical basis vectors. We set $\bm{\theta}_1^* = \bm{e}_1$ and define $\bm{\theta}_2^* = \bm{\theta}_1^* + \bm{p}$, where $\bm{p}$ represents a significant perturbation. The noise is modeled as normalized Gaussian noise. We set $t = 10$, allowing us to gather $10$ noisy projections for each bandit estimation problem, denoted by $(y_{1,s})_{s=1}^{10}$ and $(y_{2,s})_{s=1}^{10}$, with $y_{1,s} = \bm{\theta}_1^{*\top}\bm{x}_s + \eta_{1,s}$ (\resp{} $y_{2,s} = \bm{\theta}_2^{*\top} \bm{x}_s + \eta_{2,s}$). The design matrix is shared between both problems, i.e., $\bm{A}_t = \bm{A}_{1,t} = \bm{A}_{2,t} = \sum_{s=1}^{10} \bm{x}_s \bm{x}_s^\top$. At each iteration, the arm $\bm{x}_s$ is chosen at random uniformly.

\subsection{The Bandit Adaptive Sample Sharing algorithm}

We include a simplified flow chart to clarify and summarize the main steps of our algorithm.

\begin{figure}[H]
    \centering
    \includegraphics[width=.25\linewidth]{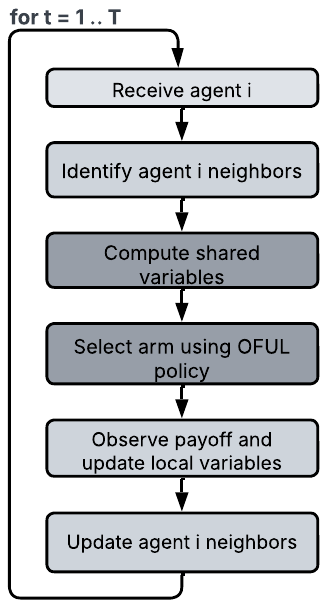}
    \caption{Overview of the Bandit Adaptive Sample Sharing (\emph{BASS}) Algorithm}
    \label{fig:alg_flowchart}
\end{figure}

In~\autoref{fig:alg_flowchart}, we summarize the main steps of the BASS algorithm, illustrating the process from selecting an arm using the OFUL policy, to identifying the agent’s neighborhood, updating shared and local variables, and updating the agent graph.

\section{Appendix: Technical lemmas} 
\label{sec:supp_technical_lemmas}

We gather here technical lemmas used later in the remaining proofs. \\

\begin{lemma}[Error vector of the difference of two estimated bandit parameters]
    \label{lm:theta_est_diff}
    
    At iteration $t$, considering the agent $i, j$ and their corresponding bandit parameter estimation $\widehat{\bm{\theta}}_{i,t}$, $\widehat{\bm{\theta}}_{i,t}$, with $\bm{\theta}_i^* \neq \bm{\theta}_j^*$, we have:
    \begin{equation*}
        \widehat{\bm{\theta}}_{i,t} - \widehat{\bm{\theta}}_{i,t} =  \bm{\theta}_i^* - \bm{\theta}_j^* +  \bm{A}_t^{-1} Z_{i,j,t} \quad \text{with } Z_{i, j} = \sum_{s=1}^{t} \bm{x}_{i, s} \eta_{i, s}  - \bm{x}_{j, s} \eta_{j, s}
    \end{equation*}
    
\end{lemma}

\begin{proof}[Proof of {\upshape\autoref{lm:theta_est_diff}}]
    
    At iteration $t$, considering the agent $i$ and their corresponding bandit parameter estimation $\widehat{\bm{\theta}}_{i,t}$ with $\bm{\theta}_i^*$ its true bandit parameter, we have:
    \begin{equation*}
       \widehat{\bm{\theta}}_{i,t} = \bm{\theta}_i^* + \bm{A}_{i,t}^{-1} \sum_{s=1}^t \bm{x}_{i, s} \eta_{i, s} \quad \text{with} \quad \bm{A}_{i,t} = \sum_{s=1}^t \bm{x}_{i, s} \bm{x}_{i, s}\top \enspace,
    \end{equation*}

    The last equation being obtained by cancelling the gradient associated with the OLS. Similarly, if we consider a second agent and their difference, we have:
    \begin{align*}
       \widehat{\bm{\theta}}_{i,t} - \widehat{\bm{\theta}}_{j,t} &= \bm{\theta}_i^* - \bm{\theta}_j^* +  \bm{A}_{i,t}^{-1} \sum_{s=1}^t \bm{x}_{i, s} \eta_{i, s} +  \bm{A}_{j,t}^{-1} \sum_{s=1}^t \bm{x}_{j, s} \eta_{j, s} \\
       &= \bm{\theta}_i^* - \bm{\theta}_j^* +  \bm{A}_t^{-1} \sum_{s=1}^t \bm{x}_{i, s} \eta_{i, s} -  \bm{x}_{j, s} \eta_{j, s} \quad \text{since synchronous pulling, we have} \quad \bm{A}_{i,t} = \bm{A}_{j,t} = \bm{A}_t \\
       &= \bm{\theta}_i^* - \bm{\theta}_j^* +  \bm{A}_t^{-1} Z_{i,j,t} \quad \text{with } Z_{i, j} = \sum_{s=1}^{t} \bm{x}_{i, s} \eta_{i, s}  - \bm{x}_{j, s} \eta_{j, s}
    \end{align*}

\end{proof}

\begin{definition}[Optimistic true bandit parameter]
    \label{def:chara_theta_opt}
    
    Let $\beta > 0$, $\widehat{\bm{\theta}} \in \mathbf{R}^d$, $\bm{x} \in  \mathbf{R}^d$ and $\bm{A} \in \mathbf{R}^{d \times d}$ a positive semi-definite matrix, we define by the \emph{optimistic true bandit parameter}, the vector $\tilde{\bm{\theta}}$ defined as:
    \begin{equation*}
        \tilde{\bm{\theta}} = \underset{\bm{\theta} \in \mathbf{R}^d}{\argmax} ~ \bm{\theta}\top\bm{x} \quad \text{such as} \quad \|\bm{\theta} - \widehat{\bm{\theta}} \|_{\bm{A}}^2 \leq \beta^2
    \end{equation*}
    
\end{definition}

\begin{lemma}[Characterization of the optimistic true bandit parameter]
    \label{lm:chara_theta_opt}
    
    Let $\beta > 0$, $\widehat{\bm{\theta}} \in \mathbf{R}^d$, $\bm{x} \in  \mathbf{R}^d$, $\bm{A} \in \mathbf{R}^{d \times d}$ a positive semi-definite matrix and $\tilde{\bm{\theta}} \in \mathbf{R}^d$ the corresponding optimistic true bandit parameter is:
    \begin{equation*}
        \tilde{\bm{\theta}} = \widehat{\bm{\theta}} + \frac{\beta}{\| \bm{x} \|_{\bm{A}^{-1}}} \bm{A}^{-1} \bm{x}
    \end{equation*}
    
\end{lemma}

\begin{proof}[Proof of {\upshape\autoref{lm:chara_theta_opt}}]

    Let consider $\beta > 0$, $\widehat{\bm{\theta}} \in \mathbf{R}^d$, $\bm{x} \in  \mathbf{R}^d$, $\bm{A} \in \mathbf{R}^{d \times d}$ a positive semi-definite matrix, the corresponding $\tilde{\bm{\theta}} \in \mathbf{R}^d$ and the following optimization problem:
    \begin{equation*}
        \tilde{\bm{\theta}} = \underset{\bm{\theta} \in \mathbf{R}^d}{\argmax} ~ \bm{\theta}\top\bm{x} \quad \text{such as} \quad \|\bm{\theta} - \widehat{\bm{\theta}} \|_{\bm{A}}^2 \leq \beta^2
    \end{equation*}

    We derive the corresponding Lagrandian: $\mathcal{L}(\bm{\theta}, \mu) = \bm{\theta}\top\bm{x} - \mu(\|\bm{\theta} - \widehat{\bm{\theta}} \|_{\bm{A}}^2 - \beta^2)$ and cancelling the associated gradient is equivalent to:

    \begin{align*}
        & \iff \begin{cases}
            \bm{A}\tilde{\bm{\theta}} = \frac{1}{2 \mu}\bm{x} + \bm{A}\widehat{\bm{\theta}} \\
            \beta^2 = \|\bm{\theta} - \widehat{\bm{\theta}} \|_{\bm{A}}^2
        \end{cases} \\
        & \iff \begin{cases}
            \tilde{\bm{\theta}} = \frac{1}{2 \mu}\bm{A}^{-1}\bm{x} + \widehat{\bm{\theta}} \\
            \| \bm{x} \|_{\bm{A}^{-1}}^2 = 4 \mu^2 \beta^2
        \end{cases} \\
        & \iff \tilde{\bm{\theta}} = \widehat{\bm{\theta}} + \frac{\beta}{\| \bm{x} \|_{\bm{A}^{-1}}}\bm{A}^{-1}\bm{x} \\
    \end{align*}
    
\end{proof}

\begin{lemma}[Positivity of the clustering error]
    \label{lm:claim}

    For each agent $i$, at probability $1-\delta$, we have:
    \begin{equation*}
        \forall t > 0 \quad \text{we have:} \quad \hat{N}_{i, t} - \rho_{m(i)} N \geq 0 \enspace.
    \end{equation*}
    
\end{lemma}

\begin{proof}[Proof of {\upshape\autoref{lm:claim}}]

    By using proof by contradiction, at iteration $t>0$, let us consider an agent $i$ and suppose that $\hat{N}_{i, t} \leq \rho_{m(i)} N$. This implies there exists an agent $j$, such as $\Psi(i, j, t) = 1$ and $\bm{\theta}_j^* = \bm{\theta}_i^*$. Let us develop $\Psi(i, j, t) = 1$, at probability $1-\delta$, we have:
    \begin{align*}
        \| \hat{\bm{\theta}}_i - \hat{\bm{\theta}}_j \|_{\bm{A}_t} &\geq 2 \beta(\delta, \bm{A}_t) \\
        \| \bm{\theta}_i^* - \bm{\theta}_j^* \|_{\bm{A}_t} + \| \bm{Z}_{i,j,t} \|_{\bm{A}^{-1}_t} &\geq 2 \beta(\delta, \bm{A}_t) \\
        \sqrt{2} \beta(\delta, \bm{A}_t) &\geq 2 \beta(\delta, \bm{A}_t) \\
    \end{align*}

    Hence, for each agent $i$, at iteration $t>0$, with probability $1-\delta$: $\hat{N}_{i, t} \geq \rho_{m(i)} N$
    
\end{proof}

\begin{lemma}[$\beta$ ratio upper-bound]
    \label{lm:beta_ratio}

    Let $\delta>0$, $N \in \mathbf{N}$, $t>0$ and $\bm{A}_t$ the design matrix obtained after $t$ iterations for a given agent, we have: 
    \begin{equation*}
        \frac{\beta(\delta, N \bm{A}_t)}{\beta(\delta, \bm{A}_t)} \leq \sqrt{1 + \frac{d \log N}{2\log \frac{1}{\delta}}}
    \end{equation*}
    
\end{lemma}

\begin{proof}[Proof of {\upshape\autoref{lm:beta_ratio}}]

    Let us consider $\delta>0$, $N \in \mathbf{N}$, $t>0$ and $\bm{A}_t$ the design matrix of a given agent, we have: 
    \begin{align*}
        \frac{\beta(\delta, N \bm{A}_t)}{\beta(\delta, \bm{A}_t)} &= \sqrt{\frac{2\log \frac{1}{\delta} + \log \mathrm{det}(N\bm{A}_t)}{2\log \frac{1}{\delta} + \log \mathrm{det}(\bm{A}_t)}} \\
         &= \sqrt{\frac{2\log \frac{1}{\delta} + d\log N + \log \mathrm{det}(\bm{A}_t)}{2\log \frac{1}{\delta} + \log \mathrm{det}(\bm{A}_t)}} \\
         &\leq \sqrt{1 + \frac{d\log N}{2\log \frac{1}{\delta}}} \quad \text{since} \quad \mathrm{det}(\bm{A}_t) \geq \mathrm{det}(\bm{A}_{t-1}) \quad \text{and} \quad \bm{A}_0 = \bm{I}
    \end{align*}
    
\end{proof}

\section{Appendix: Bandit parameter estimation error bounding} 
\label{sec:supp_single_agent_regret_minimization}

\thma*

\begin{proof}[Proof of {\upshape\autoref{th:RHS_est_theta}}]

    The proof can be founded in the supplementary material of~\citet{Abbasi2011}.

\end{proof}

\section{Appendix: Instantaneous regret upper bounds} 
\label{sec:supp_instantaneous_regret_ub}

\lema*

\begin{proof}[Proof of {\upshape\autoref{lem:regret_bound_inst}}]

For the single case, at round $t$, for a given agent $i$, by considering the optimistic arm $\bm{x}_{i,t}$ and the optimal arm $\bm{x}_{i}^*$, we have:

\begin{align*}
    r_{i, t} &= \bm{\theta}_i^{* \top} (\bm{x}_i^* - \bm{x}_{i,t}) \\
    &= \bm{\theta}^*_i\top\bm{x}_i^* - \bm{\theta}^*_i\top\bm{x}_{i,t} \\
    &\leq \tilde{\bm{\theta}}_{i,t}^\top \bm{x}_{i,t} - \bm{\theta}^*_i\top\bm{x}_{i,t} \quad \text{since} ~ (\tilde{\bm{\theta}}_{i,t}, \bm{x}_{i,t}) \quad \text{are the optimistic tuple} \\
    &\leq \left(\widehat{\bm{\theta}}_{i,t} + \frac{\beta(\delta, \bm{A}_t)}{\|\bm{x}_{i,t}\|_{\bm{A}_t^{-1}}} \bm{A}_t^{-1}\bm{x}_{i,t} - \bm{\theta}^*_i\right)^\top\bm{x}_{i,t} \quad \text{from~\autoref{lm:chara_theta_opt}} \\
    &\leq \left\|\widehat{\bm{\theta}}_{i,t} - \bm{\theta}^*_i\right\|_{\bm{A}_t} \left\|\bm{x}_{i,t}\right\|_{\bm{A}_t^{-1}} + \frac{\beta(\delta, \bm{A}_t)}{\|\bm{x}_{i,t}\|_{\bm{A}_t^{-1}}} \|\bm{x}_{i,t}\|^2_{\bm{A}_t^{-1}} \\
    &\leq 2 \beta(\delta, \bm{A}_t) \|\bm{x}_{i,t}\|_{\bm{A}_t^{-1}}
\end{align*}

Moreover, for the collaborative case, at round $t$, for a given agent $i$, by considering the optimistic arm $\bm{x}_{\widehat{\mathcal{N}}_{i}(t)}$ and the optimal arm $\bm{x}_{i}^*$, we have:

\begin{align}
    r_{i, t} &= \bm{\theta}_i^{* \top} (\bm{x}_i^* - \bm{x}_{\widehat{\mathcal{N}}_{i}(t)}) \notag\\
    &= (\bm{\theta}^*_{\widehat{\mathcal{N}}_{i}(t)} - \bm{\theta}^*_{\widehat{\mathcal{N}}_{i}(t)} + \bm{\theta}_i^*)^\top (\bm{x}_i^* - \bm{x}_{\widehat{\mathcal{N}}_{i}(t)}) \notag\\
    &= \bm{\theta}_{\widehat{\mathcal{N}}_{i}(t)}^{* \top}\bm{x}_i^* - \bm{\theta}_{\widehat{\mathcal{N}}_{i}(t)}^{* \top}\bm{x}_{\widehat{\mathcal{N}}_{i}(t)} +  (\bm{\theta}_i^* - \bm{\theta}_{\widehat{\mathcal{N}}_{i}(t)}^*)^\top(\bm{x}_i^* - \bm{x}_{\widehat{\mathcal{N}}_{i}(t)})  \notag\\
    &\leq \tilde{\bm{\theta}}_{\widehat{\mathcal{N}}_{i}(t)}^\top\bm{x}_{\widehat{\mathcal{N}}_{i}(t)} - \bm{\theta}_{\widehat{\mathcal{N}}_{i}(t)}^{* \top}\bm{x}_{\widehat{\mathcal{N}}_{i}(t)} +  (\bm{\theta}_i^* - \bm{\theta}^*_{\widehat{\mathcal{N}}_{i}(t)})^\top(\bm{x}_i^* - \bm{x}_{\widehat{\mathcal{N}}_{i}(t)}) \quad \text{since} ~ (\tilde{\bm{\theta}}_{\widehat{\mathcal{N}}_{i}(t)}, \bm{x}_{\widehat{\mathcal{N}}_{i}(t)}) \quad \text{are the optimistic tuple} \notag\\
    &\leq (\tilde{\bm{\theta}}_{\widehat{\mathcal{N}}_{i}(t)} - \bm{\theta}^*_{\widehat{\mathcal{N}}_{i}(t)})^\top\bm{x}_{\widehat{\mathcal{N}}_{i}(t)} +  (\bm{\theta}_i^* - \bm{\theta}^*_{\widehat{\mathcal{N}}_{i}(t)})^\top(\bm{x}_i^* - \bm{x}_{\widehat{\mathcal{N}}_{i}(t)}) \label{eq:instantaneous_regret_beginning}
\end{align}

We now consider the corresponding norms, we have:
\begin{align*}
    &\leq (\tilde{\bm{\theta}}_{\widehat{\mathcal{N}}_{i}(t)} - \bm{\theta}^*_{\widehat{\mathcal{N}}_{i}(t)})^\top\bm{x}_{\widehat{\mathcal{N}}_{i}(t)} +  \|\bm{\theta}_i^* - \bm{\theta}^*_{\widehat{\mathcal{N}}_{i}(t)}\|_{\bm{A}_t}\|\bm{x}_i^* - \bm{x}_{\widehat{\mathcal{N}}_{i}(t)}\|_{\bm{A}_t^{-1}} \\
    &\leq (\tilde{\bm{\theta}}_{\widehat{\mathcal{N}}_{i}(t)} - \bm{\theta}^*_{\widehat{\mathcal{N}}_{i}(t)})^\top\bm{x}_{\widehat{\mathcal{N}}_{i}(t)} + 2 \|\bm{\theta}^*_{\widehat{\mathcal{N}}_{i}(t)} - \bm{\theta}_i^*\|_{\bm{A}_t} \quad \text{from~\autoref{assum:bounded_norms}} \\
    &\leq \left(\widehat{\bm{\theta}}_{\widehat{\mathcal{N}}_{i}(t)} + \frac{\beta(\delta, \bm{A}_{\widehat{\mathcal{N}}_{i}(t)})}{\|\bm{x}_{\widehat{\mathcal{N}}_{i}(t)}\|_{\bm{A}_{\widehat{\mathcal{N}}_{i}(t)}^{-1}}} \bm{A}_{\widehat{\mathcal{N}}_{i}(t)}^{-1 \top}\bm{x}_{\widehat{\mathcal{N}}_{i}(t)} - \bm{\theta}^*_{\widehat{\mathcal{N}}_{i}(t)}\right)^\top\bm{x}_{\widehat{\mathcal{N}}_{i}(t)} + 2 \|\bm{\theta}^*_{\widehat{\mathcal{N}}_{i}(t)} - \bm{\theta}_i^*\|_{\bm{A}_t} \quad \text{from~\autoref{lm:chara_theta_opt}} \\
    &\leq \left(\frac{\beta(\delta, \bm{A}_{\widehat{\mathcal{N}}_{i}(t)})}{\|\bm{x}_{\widehat{\mathcal{N}}_{i}(t)}\|_{\bm{A}_{\widehat{\mathcal{N}}_{i}(t)}^{-1}}} \bm{A}_{\widehat{\mathcal{N}}_{i}(t)}^{-1 \top}\bm{x}_{\widehat{\mathcal{N}}_{i}(t)}\right)^\top\bm{x}_{\widehat{\mathcal{N}}_{i}(t)} + \left( \widehat{\bm{\theta}}_{\widehat{\mathcal{N}}_{i}(t)} - \bm{\theta}^*_{\widehat{\mathcal{N}}_{i}(t)} \right)^\top\bm{x}_{\widehat{\mathcal{N}}_{i}(t)} + 2 \|\bm{\theta}^*_{\widehat{\mathcal{N}}_{i}(t)} - \bm{\theta}_i^*\|_{\bm{A}_t} \\
    &\leq \beta(\delta, \bm{A}_{\widehat{\mathcal{N}}_{i}(t)}) \|\bm{x}_{\widehat{\mathcal{N}}_{i}(t)}\|_{\bm{A}_{\widehat{\mathcal{N}}_{i}(t)}^{-1}} + \|\widehat{\bm{\theta}}_{\widehat{\mathcal{N}}_{i}(t)} - \bm{\theta}^*_{\widehat{\mathcal{N}}_{i}(t)}\|_{\bm{A}_{\widehat{\mathcal{N}}_{i}(t)}} \|\bm{x}_{\widehat{\mathcal{N}}_{i}(t)}\|_{\bm{A}_{\widehat{\mathcal{N}}_{i}(t)}^{-1}} + 2 \|\bm{\theta}^*_{\widehat{\mathcal{N}}_{i}(t)} - \bm{\theta}_i^*\|_{\bm{A}_t} \\
    &\leq \left(\beta(\delta, \bm{A}_{\widehat{\mathcal{N}}_{i}(t)}) + \|\widehat{\bm{\theta}}_{\widehat{\mathcal{N}}_{i}(t)} - \bm{\theta}^*_{\widehat{\mathcal{N}}_{i}(t)}\|_{\bm{A}_{\widehat{\mathcal{N}}_{i}(t)}}\right) \|\bm{x}_{\widehat{\mathcal{N}}_{i}(t)}\|_{\bm{A}_{\widehat{\mathcal{N}}_{i}(t)}^{-1}} + 2 \|\bm{\theta}^*_{\widehat{\mathcal{N}}_{i}(t)} - \bm{\theta}_i^*\|_{\bm{A}_t} \\
    &\leq 2\beta(\delta, \bm{A}_{\widehat{\mathcal{N}}_{i}(t)})\|\bm{x}_{\widehat{\mathcal{N}}_{i}(t)}\|_{\bm{A}_{\widehat{\mathcal{N}}_{i}(t)}^{-1}} + 2 \|\bm{\theta}^*_{\widehat{\mathcal{N}}_{i}(t)} - \bm{\theta}_i^*\|_{\bm{A}_t} \quad \text{by definition of} \quad \beta(\delta, \bm{A}_{\widehat{\mathcal{N}}_{i}(t)}) \\
    &\leq \frac{2}{\sqrt{N}}\beta(\delta, N \bm{A}_t)\|\bm{x}_{\widehat{\mathcal{N}}_{i}(t)}\|_{\bm{A}_t^{-1}} + \frac{2}{N} \sum_{j=1}^N \|\bm{\theta}_j^* - \bm{\theta}_i^*\|_{\bm{A}_t}
\end{align*}

\end{proof}

\section{Appendix: Collaboration condition} 
\label{sec:supp_initial_sep_criteria}

We proof Equation~\autoref{eq:stop_crit}. With~\autoref{assum:bounded_norms}, we can bound the term $\|\bm{x}_{i,t}\|_{\bm{A}_t^{-1}}$ in~\autoref{lem:regret_bound_inst}, leading to:
\begin{equation}
    \label{eq:ub_inst_regret_single}
    r_{i, t} \leq 2 \beta(\delta, \bm{A}_t)
\end{equation}

 Similarly, for the collaborating case of two agents, we have:
\begin{equation}
    \label{eq:ub_inst_regret_collab}
    r_{i, t} \leq \sqrt{2}\beta(\delta, 2 \bm{A}_t) + \|\bm{\theta}_j^* - \bm{\theta}_i^*\|_{\bm{A}_t}
\end{equation}

Equation~\autoref{eq:ub_inst_regret_single} and Equation~\autoref{eq:ub_inst_regret_collab} leads us to:
\begin{align*}
    \frac{1}{\sqrt{2}} \frac{\beta(\delta, 2 \bm{A}_t)}{\beta(\delta, \bm{A}_t)} + \frac{\|\bm{\theta}_j^* - \bm{\theta}_i^*\|_{\bm{A}_t}}{2 \beta(\delta, \bm{A}_t)} &\leq 1 \\
     \frac{\|\bm{\theta}_j^* - \bm{\theta}_i^*\|_{\bm{A}_t}}{2 \beta(\delta, \bm{A}_t)} &\leq 1 -\frac{1}{\sqrt{2}} \frac{\beta(\delta, 2 \bm{A}_t)}{\beta(\delta, \bm{A}_t)} \\
     \frac{\|\bm{\theta}_j^* - \bm{\theta}_i^*\|_{\bm{A}_t}}{2 \beta(\delta, \bm{A}_t)} &\leq 1 -\frac{1}{\sqrt{2}} \\
     \|\bm{\theta}_j^* - \bm{\theta}_i^*\|_{\bm{A}_t} &\leq (2 - \sqrt{2}) \beta(\delta, \bm{A}_t)
\end{align*}

\section{Appendix: Ellipsoid separation under synchronous pulling} 
\label{sec:supp_gili_equivalence}

\lemb*

\begin{proof}[Proof of {\upshape\autoref{cor:ell_test}}]

    The proof can be founded in the supplementary material of~\citet{Gilitschenski2012}.

\end{proof}
\section{Appendix: Ellipsoid separation test function} 
\label{sec:supp_our_test_equivalence}

We prove the ellipsoid separation property of the $\Psi$ function.

\begin{restatable}[$\kappa$ellipsoid separation property of the $\Psi$ function.]{property}{thmh}
    \label{th:gili_ell_test_equivalent}

    Let $0 < \delta < 1$, for two agents $i$ and $j$, and following~\autoref{assum:bounded_norms} and~\autoref{assum:sub_gaussian_noise}, the following statements are equivalent:
    \begin{equation*}
        \forall ~ i, j, t>0 \quad \Psi(i, j, t) = 1 \iff \mathcal{E}(\hat{\bm{\theta}}_i, \bm{A}_t, \tilde{\beta}) \cap \mathcal{E}(\hat{\bm{\theta}}_j, \bm{A}_t, \tilde{\beta}) = \emptyset \enspace,
    \end{equation*}

\end{restatable}

\begin{proof}[Proof of {\upshape\autoref{th:gili_ell_test_equivalent}}]

 Let us consider two agents $i$ and $j$, with their associated unknown bandit parameter estimates $\hat{\bm{\theta}}_i$ and $\hat{\bm{\theta}}_j$, the observation matrices $\bm{A}_i$ and $\bm{A}_j$ along with their local confidence level briefly noted $\beta_i$ and $\beta_j$ as in~\eqref{th:RHS_est_theta}. If we consider that $\bm{\Phi}$ and $\mathrm{diag}(\bm{\eta})$, the eigenvectors and the eigenvalues of the generalized eigenvalue problem, we have:
\begin{equation*}
    \bm{A}_i \bm{\Phi} = \bm{A}_j \bm{\Phi} \mathrm{diag}(\bm{\eta}) \quad \text{with} ~ \bm{\Phi}_i^\top \bm{A}_j \bm{\Phi}_i = 1
\end{equation*}    

First, with the given scaling of this problem being, one can recover the following equivalent constraints:
\begin{equation*} 
    \begin{cases}
        \bm{\Phi}^\top \bm{A}_i \bm{\Phi} = \mathrm{diag}(\bm{\eta}) \\
        \bm{\Phi}^\top \bm{A}_j \bm{\Phi} = \bm{I} \\
    \end{cases}
\end{equation*} 

Which gives us:
\begin{equation} 
    \label{eq:Ai_Aj}
    \begin{cases}
        \bm{A}_i^{-1} = \bm{\Phi}^\top \mathrm{diag}(\bm{\eta}) \bm{\Phi} \\
        \bm{A}_j^{-1} = \bm{\Phi}^\top \bm{\Phi} \\
    \end{cases}
\end{equation} 

Secondly, if we consider the function $\forall s \in ]0, 1[ \quad \psi_{i,j}(s) = \frac{\gamma^2}{4} - \sum_{l=1}^d \mu_l^2 \frac{s (1 - s)}{\beta_i^2 + s (\beta_j^2\eta_l - \beta_i^2)}$ and $\bm{\mu} = \bm{\Phi}^\top (\hat{\bm{\theta}}_i - \hat{\bm{\theta}}_j)$ from~\autoref{def:ell_test}, we have the following:

\begin{align*}
    \psi_{i,j}(s) &= \frac{\gamma^2}{4} - \sum_{l=1}^d \mu_l^2 \frac{s (1 - s)}{\beta_i^2 + s (\beta_j^2\eta_l - \beta_i^2)} \\
    &= \frac{\gamma^2}{4} - \bm{\mu}^\top \mathrm{diag}(\frac{s(1-s)}{\beta_i^2 + s(\beta_j^2\eta_l - \beta_i^2)}) \bm{\mu} \\
    &= \frac{\gamma^2}{4} -  \bm{\mu}^\top \mathrm{diag}(\frac{\beta_i^2}{s} + \frac{\beta_j^2}{1-s}\eta_l)^{-1} \bm{\mu} \\
    &= \frac{\gamma^2}{4} -  \bm{\mu}^\top \left( \frac{\beta_i^2}{s}\bm{I} + \frac{\beta_j^2}{1-s} \mathrm{diag}(\bm{\eta}) \right)^{-1} \bm{\mu} \\
    &= \frac{\gamma^2}{4} - (\hat{\bm{\theta}}_i - \hat{\bm{\theta}}_j)^\top \left( \frac{\beta_i^2}{s}\bm{\Phi}\bm{\Phi}^\top + \frac{\beta_j^2}{1-s} \bm{\Phi}\mathrm{diag}(\bm{\eta})\bm{\Phi}^\top \right)^{-1}(\hat{\bm{\theta}}_i - \hat{\bm{\theta}}_j) \\
    &= \frac{\gamma^2}{4} - (\hat{\bm{\theta}}_i - \hat{\bm{\theta}}_j)^\top \left( \frac{\beta_i^2}{s}\bm{A}_j^{-1} + \frac{\beta_j^2}{1-s} \bm{A}_i^{-1} \right)^{-1}(\hat{\bm{\theta}}_i - \hat{\bm{\theta}}_j) \quad \text{from Equation~\autoref{eq:Ai_Aj}}\\
    &= \frac{\gamma^2}{4} - \| \hat{\bm{\theta}}_i - \hat{\bm{\theta}}_j \|_{\left(\frac{\beta_i^2}{1-s} \bm{A}_{i}^{-1} + \frac{\beta_j^2}{s} \bm{A}_{j}^{-1} \right)^{-1}}^2
\end{align*}

For $\gamma=2$, we recover the function $\kappa$ as defined in Equation~\autoref{eq:gili_func}, from~\citet{Gilitschenski2012}, and conclude the property.

\end{proof}

\section{Appendix: Lower bound on the separation time} 
\label{sec:supp_Ts_lb}

With our algorithm, once two agents are separated at iteration $t = T_s(i, j)$ (\ie{} $\Psi(i, j, t) = 1$), they no longer share samples. Hence the separation is definitive. We proof the lower bound on the separation time based on the parameter gap.

\thmb*

\begin{proof}[Proof of {\upshape\autoref{th:lb_ts}}]

To derive a lower bound, we examine a simplified case where the separation time occurs earlier than in any typical scenario.

We consider two agents $i$ and $j$, an iteration $T_s(i, j) \leq t$, we assume the nature of the noise is known and modeled as normalized Gaussian noise. We suppose that the set of arm is reduce to a single arm, \ie{} $\mathcal{X} = \left\{ \frac{\bm{\theta}_i^* - \bm{\theta}_j^*}{\| \bm{\theta}_i^* - \bm{\theta}_j^* \|_2} \right\}$, which the most favorable case to distinguish the two bandit parameters.

First, we bound $\| \hat{\bm{\theta}}_t - \bm{\theta}_i^* \|_{\bm{A}_t}$ in our specific setting:
\begin{align*}
    \| \hat{\bm{\theta}}_t - \bm{\theta}_i^* \|_{\bm{A}_t} &= \| \bm{A}_t^{\frac{1}{2}} ( \hat{\bm{\theta}}_t - \bm{\theta}_i^*) \|_2 \\
    &= \| \frac{1}{\sqrt{t}} \bm{A}_t ( \hat{\bm{\theta}}_t - \bm{\theta}_i^*) \|_2 \quad \text{since} \quad \bm{A}_t^{\frac{1}{2}} = \frac{1}{\sqrt{t}} \bm{A}_t = \sqrt{t} \bm{x} \bm{x}^\top \\
    &= \left| \frac{1}{\sqrt{t}} \sum_{s=1}^t \eta_s \right| \quad \text{with} \quad \frac{1}{\sqrt{t}} \sum_{s=1}^t \eta_s \sim \mathcal{N}(0,1) \\
\end{align*}

We have $\mathbf{P}\left[ \frac{1}{\sqrt{t}} \sum_{s=1}^t \eta_s \geq a \right] = 1 - \mathrm{erf}(\frac{a}{\sqrt{2}})$. So with probability $1 - \delta$, we have:
\begin{equation*}
    \| \hat{\bm{\theta}}_t - \bm{\theta}_i^* \|_{\bm{A}_t} \leq \beta(\delta) \quad \text{with} \quad \beta(\delta) = \sqrt{2} \mathrm{erf}^{-1}(1-\delta) \enspace.
\end{equation*}

Secondly, we bound $\| \hat{\bm{\theta}}_{i,t} - \hat{\bm{\theta}}_{j,t} \|_{\bm{A}_t}$ in our setting similarly to the previous approach, we have:
\begin{align*}
    \| \hat{\bm{\theta}}_{i,t} - \hat{\bm{\theta}}_{j,t} \|_{\bm{A}_t} &= \left\| \left( \sqrt{t} \| \bm{\theta}_i^* - \bm{\theta}_j^* \|_2  + \frac{1}{\sqrt{t}} \sum_{s=1}^t (\eta_{i,s} - \eta_{j,s}) \right) \bm{x} \right\|_2 \\
    &= |Z_{i,j,t}| \quad \text{with} \quad Z_{i,j,t} \sim \mathcal{N}(\sqrt{t} \| \bm{\theta}_i^* - \bm{\theta}_j^* \|_2, 2) \\
\end{align*}

Lastly we consider our collaboration criterion with $t = T_s(i,j)$, we have:
\begin{align*}
    \| \hat{\bm{\theta}}_{i,t} - \hat{\bm{\theta}}_{j,t} \|_{\bm{A}_t} &\geq 2 \beta(\delta) \\
    |Z_{i,j,t}| &\geq 2 \beta(\delta)
\end{align*}

Moreover setting $\mathbf{P}\left[ |Z_{i,j,t}| \geq 2 \beta(\delta) \right] = 1 - \delta$ and since $\mathbf{P}\left[ |Z_{i,j,t}| \geq 2 \beta(\delta) \right] = 2 (1 - \mathbf{P}\left[Z_{i,j,t} \leq 2 \beta(\delta) \right])$, we have:
\begin{align*}
   1 - \delta &= 2 \left(1 - \left(\frac{1}{2} + \frac{1}{2} \mathrm{erf}\left(\frac{1}{2\sqrt{2}}\left(2 \beta(\delta) - \sqrt{t} \| \bm{\theta}_i^* - \bm{\theta}_j^* \|_2\right)\right)\right)\right) \\
   1 - \delta &= 1 - \mathrm{erf}\left(\frac{1}{2\sqrt{2}}\left(2 \beta(\delta) - \sqrt{t} \| \bm{\theta}_i^* - \bm{\theta}_j^* \|_2\right)\right) \\
   2 \sqrt{2} \mathrm{erf}^{-1}(\delta) &= 2 \beta(\delta) - \sqrt{t} \| \bm{\theta}_i^* - \bm{\theta}_j^* \|_2
\end{align*}

So, at iteration $t = T_s(i,j) = \frac{4(\beta(\delta) - \sqrt{2}\mathrm{erf}^{-1}(\delta))^2}{\| \bm{\theta}_i^* - \bm{\theta}_j^* \|_2}$ with probability $1-\delta$, we have $\| \hat{\bm{\theta}}_{i,t} - \hat{\bm{\theta}}_{j,t} \|_{\bm{A}_t} \geq 2 \beta(\delta)$, which leads to:
\begin{equation*}
    T_s(i,j) \geq \left\lceil \frac{8(\mathrm{erf}^{-1}(1-\delta) - \mathrm{erf}(\delta))^2}{\| \bm{\theta}_i^* - \bm{\theta}_j^* \|_2} \right\rceil
\end{equation*}

Although we focus on a simple case, we demonstrate that the separation time scales inversely with $\| \bm{\theta}_i^* - \bm{\theta}_j^* \|_2$ which characterize how difficult the problem is.

\end{proof}

\section{Appendix: Upper bound on the separation time} 
\label{sec:supp_Ts_ub}

\thmc*

\begin{proof}[Proof of {\upshape\autoref{th:ub_ts}}]

Given two agents $i$ and $j$, at iteration $t \leq T_s(i, j)$, from~\citet{Gilitschenski2012}, under a synchronous pulling, we have:
\begin{align*}
    \| \hat{\bm{\theta}}_{i,t} - \hat{\bm{\theta}}_{j,t} \|_{\bm{A}_t} &\leq 2 \beta(\delta, \bm{A}_t) \\
    \| \bm{\theta}_i^* - \bm{\theta}_j^* \|_{\bm{A}_t} - \| \bm{Z}_{i,j,t} \|_{\bm{A}_t^{-1}} &\leq 2 \beta(\delta, \bm{A}_t) \quad \text{from~\autoref{lm:theta_est_diff}} \\
    \| \bm{\theta}_i^* - \bm{\theta}_j^* \|^2_{\bm{A}_t} &\leq (2 + \sqrt{2})^2 \beta(\delta, \bm{A}_t)^2 \\
    \sum_{k=1}^K T_{kt} (\bm{x}_k^\top (\bm{\theta}_i^* - \bm{\theta}_j^*) )^2 &\leq (2 + \sqrt{2})^2 \beta(\delta, \bm{A}_t)^2  \quad \text{with $T_{kt}$ the number of pulling of arm $\bm{x}_k$ at iteration $t$} \\
\end{align*}

We have $T_{kt} = \epsilon_e T_{kt}^{\mathrm{Unif}} + (1 - \epsilon_e) T_{kt}^{\mathrm{UCB}}$, with $T_{kt}^{\mathrm{Unif}}$ the number of pulling of arm $\bm{x}_k$ at iteration $t$ following the \emph{Uniform} policy\footnote{This highlight the particular utility of the $\epsilon-$uniform sampling , as the typical UCB lower bounds are asymptotically.} and $T_{kt}^{\mathrm{UCB}}$ the number of pulling of arm $\bm{x}_k$ at iteration $t$ following the \emph{UCB} policy as depicted in~\autoref{alg:bass}, we have with $t$ large enough:
\begin{equation*}
    T_{kt} = \epsilon_e T_{kt}^{\mathrm{Unif}} + (1 - \epsilon_e) T_{kt}^{\mathrm{UCB}} \geq \epsilon_e T_{kt}^{\mathrm{Unif}} = \epsilon_e \frac{t}{K}
\end{equation*}

Thus, we have:
\begin{align*}
    \epsilon_e \frac{t}{K} \sum_{k=1}^K (\bm{x}_k^\top (\bm{\theta}_i^* - \bm{\theta}_j^*) )^2 &\leq (2 + \sqrt{2})^2 \beta(\delta, \bm{A}_t)^2 \\
    \epsilon_e \frac{t}{K} \sum_{k=1}^K (\bm{x}_k^\top (\bm{\theta}_i^* - \bm{\theta}_j^*) )^2 &\leq (2 + \sqrt{2})^2 R^2 \left(2 \log \frac{1}{\delta} + \log \mathrm{det}(\bm{A}_t)\right) \\
    \epsilon_e \frac{t}{K} \sum_{k=1}^K (\bm{x}_k^\top (\bm{\theta}_i^* - \bm{\theta}_j^*) )^2 &\leq (2 + \sqrt{2})^2 R^2 \left(2 \log \frac{1}{\delta} + \frac{d}{2} \log \left(1 + \frac{t}{d} \right) \right) \\
    \epsilon_e \frac{t}{K} \sum_{k=1}^K (\bm{x}_k^\top (\bm{\theta}_i^* - \bm{\theta}_j^*) )^2 &\leq (2 + \sqrt{2})^2 R^2 \left(2 \log \frac{1}{\delta} + \frac{t}{2}\right) \\
    t\left( \frac{\epsilon_e}{K} \sum_{k=1}^K (\bm{x}_k^\top (\bm{\theta}_i^* - \bm{\theta}_j^*) )^2 - \frac{(2 + \sqrt{2})^2 R^2}{2}\right) &\leq 2 (2 + \sqrt{2})^2 R^2 \log \frac{1}{\delta} \\
\end{align*}

Finally, we conclude that:
\begin{equation*}
    T_s(i, j) \leq \left\lceil \frac{4 (2 + \sqrt{2})^2 R^2 \log \frac{1}{\delta}}{2 \epsilon_e Q_{i,j} - (2 + \sqrt{2})^2 R^2} \right\rceil
\end{equation*}
    with an Signal to Noise Ratio (SNR) such as: $\frac{\sqrt{Q_{i,j}}}{R} \geq \frac{(2 + \sqrt{2})}{\sqrt{2 \epsilon_e}}$ and $Q_{i,j} = \frac{1}{K} \sum_{k=1}^{K}(\bm{x}_k^\top(\bm{\theta}_i^* - \bm{\theta}_j^*))^2$

\end{proof}

\section{Appendix: Cumulative pseudo-regret upper bound} 
\label{sec:supp_first_cumul_regret_ub}

\thmd*

\begin{proof}[Proof of {\upshape\autoref{th:upper_bound_general}}]

We consider the case of two agents $i$ and $j$, focusing the instantaneous regret $r_{i, t}$, given that $t\leq T_s(i,j)$ we have $\mathcal{N}_{i}(t) = \{j\}$, from Equation~\autoref{eq:instantaneous_regret_beginning}, we have:

\begin{align*}
    r_{i, t} &\leq \left( \tilde{\bm{\theta}}_{\widehat{\mathcal{N}}_{i}(t)} - \bm{\theta}_{\widehat{\mathcal{N}}_{i}(t)}^* - \bm{\theta}_i^* + \bm{\theta}^*_{\widehat{\mathcal{N}}_{i}(t)} \right)^\top \bm{x}_{\widehat{\mathcal{N}}_{i}(t)} + (\bm{\theta}_i^* - \bm{\theta}^*_{\widehat{\mathcal{N}}_{i}(t)})^\top \bm{x}_i^* \\
    &\leq \left( \frac{\beta(\delta, \bm{A}_{\widehat{\mathcal{N}}_{i}(t)})}{\|\bm{x}_{\widehat{\mathcal{N}}_{i}(t)}\|_{\bm{A}_{\widehat{\mathcal{N}}_{i}(t)}^{-1}}} \bm{A}_{\widehat{\mathcal{N}}_{i}(t)}^{-1 \top}\bm{x}_{\widehat{\mathcal{N}}_{i}(t)} + \widehat{\bm{\theta}}_{\widehat{\mathcal{N}}_{i}(t)} - \bm{\theta}_i^* \right)^\top \bm{x}_{\widehat{\mathcal{N}}_{i}(t)} + \frac{1}{2}(\bm{\theta}_i^* - \bm{\theta}^*_j)^\top \bm{x}_i^* \\
    &\leq \beta(\delta, \bm{A}_{\widehat{\mathcal{N}}_{i}(t)}) \|\bm{x}_{\widehat{\mathcal{N}}_{i}(t)}\|_{\bm{A}_{\widehat{\mathcal{N}}_{i}(t)}^{-1}} + \left( \frac{1}{2}(\widehat{\bm{\theta}}_i + \widehat{\bm{\theta}}_j) - \bm{\theta}_i^* + \frac{1}{2}(\widehat{\bm{\theta}}_i - \widehat{\bm{\theta}}_i) \right)^\top \bm{x}_{\widehat{\mathcal{N}}_{i}(t)} + \frac{1}{2}(\bm{\theta}_i^* - \bm{\theta}^*_j)^\top \bm{x}_i^* \\
    &\leq \frac{1}{\sqrt{2}} \beta(\delta, 2 \bm{A}_t) \|\bm{x}_{\widehat{\mathcal{N}}_{i}(t)}\|_{\bm{A}_t^{-1}} + \left( \frac{1}{2} \| \widehat{\bm{\theta}}_i - \widehat{\bm{\theta}}_j \|_{\bm{A}_t} + \| \widehat{\bm{\theta}}_i - \bm{\theta}_i^*\|_{\bm{A}_t} \right) \| \bm{x}_{\widehat{\mathcal{N}}_{i}(t)} \|_{\bm{A}_t^{-1}} + \frac{1}{2}(\bm{\theta}_i^* - \bm{\theta}^*_j)^\top \bm{x}_i^* \\
    &\leq \left( \frac{1}{\sqrt{2}} \beta(\delta, 2 \bm{A}_t) + \frac{\gamma}{2} \beta(\delta, \bm{A}_t) + \beta(\delta, \bm{A}_t) \right) \| \bm{x}_{\widehat{\mathcal{N}}_{i}(t)} \|_{\bm{A}_t^{-1}} + \frac{1}{2}(\bm{\theta}_i^* - \bm{\theta}^*_j)^\top \bm{x}_i^* \\
    &\leq 2 \left( \frac{1}{2\sqrt{2}} \frac{\beta(\delta, 2 \bm{A}_t)}{\beta(\delta, \bm{A}_t)} + \frac{\gamma}{4}  + \frac{1}{2} \right) \beta(\delta, \bm{A}_t) \| \bm{x}_{\widehat{\mathcal{N}}_{i}(t)} \|_{\bm{A}_t^{-1}} + \frac{1}{2}(\bm{\theta}_i^* - \bm{\theta}^*_j)^\top \bm{x}_i^* \\
    &\leq 2 \left( \frac{1}{2} + \frac{\gamma}{4} + \frac{1}{2\sqrt{2}} \frac{\beta(\delta, 2 \bm{A}_t)}{\beta(\delta, \bm{A}_t)} \right) \beta(\delta, \bm{A}_t) \| \bm{x}_{\widehat{\mathcal{N}}_{i}(t)} \|_{\bm{A}_t^{-1}} + \frac{1}{2}(\bm{\theta}_i^* - \bm{\theta}^*_j)^\top \bm{x}_i^* \\
    &\leq 2 \left( \frac{1}{2} + \frac{\gamma}{4} + \frac{1}{2\sqrt{2}} \sqrt{1 + \frac{d \log 2}{2\log \frac{1}{\delta}}} \right) \beta(\delta, \bm{A}_t) \| \bm{x}_{\widehat{\mathcal{N}}_{i}(t)} \|_{\bm{A}_t^{-1}} + \frac{1}{2}(\bm{\theta}_i^* - \bm{\theta}^*_j)^\top \bm{x}_i^* \quad \text{from~\autoref{lm:beta_ratio}} \\
    &\leq 2 \mu(\delta, d, \gamma) \beta(\delta, \bm{A}_t) \| \bm{x}_{\widehat{\mathcal{N}}_{i}(t)} \|_{\bm{A}_t^{-1}} + \frac{1}{2}(\bm{\theta}_i^* - \bm{\theta}^*_j)^\top \bm{x}_i^*  \quad \text{with} \quad \mu(\delta, d, \gamma) = \frac{1}{2} + \frac{\gamma}{4} + \frac{1}{2\sqrt{2}} \sqrt{1 + \frac{d \log 2}{2\log \frac{1}{\delta}}} \\
\end{align*}

With~\citet{Abbasi2011} and by setting $\nu(\delta, T_s) =  \sqrt{ 4 \beta(\delta, \bm{A}_{T_s})^2 T_s d \log (1 + \frac{T_s}{d} ) }$, we have:
\begin{equation*}
    R_{i, 0, T_s}^{\mathrm{collab}} \leq \mu(\delta, d, \gamma) \nu(\delta, T_s) + \frac{T_s}{2} (\bm{\theta}_i^* - \bm{\theta}_j^*)^\top \bm{x}_i^* \enspace.
\end{equation*}

\end{proof}

Note that in order to enforce $\mu(\delta, d, \gamma) < 1$, we need to set $\gamma < 2 - \sqrt{2}\sqrt{1 + (d \log 2) / (2 \log \frac{1}{\delta})}$. For example, for $d=10$ and $\delta=0.001$, we have is $\gamma \leq 0.27$.

Moreover, recall that in~\citet{Gilitschenski2012}, the authors consider $\gamma = 2$ to derive their ellipsoid separation test. Here, the term $\sqrt{1 + (d \log 2) / (2 \log \frac{1}{\delta})}$ could be interpreted as a correction that accounts for the regret minimization objective. Note that this term is derived from the loose upper-bound of $\beta(\delta, 2\bm{A}_t) / \beta(\delta, \bm{A}_t)$, which is close to $1$ and relatively insensitive to both the number of agents and the problem's dimensionality.

\section{Appendix: Network cumulative pseudo-regret upper bound} 
\label{sec:supp_second_cumul_regret_ub}

\thme*

\begin{proof}[Proof of {\upshape\autoref{th:upper_bound_averaged_general}}]

We consider the case of two agents $i$ and $j$, focusing the averaged instantaneous regret $\bar{r}_{i, t} = \frac{1}{2}(r_{i, t} + r_{j, t})$, given that $t\leq T_s(i,j)$ we have $\mathcal{N}_{i}(t) = \{j\}$ and $\mathcal{N}_{j}(t) = \{i\}$, which give us:

\begin{align*}
    \bar{r}_{i, t} &= \frac{1}{2}(r_{i, t} + r_{j, t}) \\
    &= \frac{1}{2}(\bm{\theta}_i^{* \top} (\bm{x}_i^* - \bm{x}_{\widehat{\mathcal{N}}_{i}(t)}) + \bm{\theta}_j^{* \top} (\bm{x}_j^* - \bm{x}_{\widehat{\mathcal{N}}_{i}(t)}))\\
    &= - \bm{\theta}_{\widehat{\mathcal{N}}_{i}(t)}^{* \top}\bm{x}_{\widehat{\mathcal{N}}_{i}(t)} + \frac{1}{2} \bm{\theta}_i^{* \top}\bm{x}_i^* + \frac{1}{2} \bm{\theta}_j^{* \top}\bm{x}_j^* \\
    &= \bm{\theta}_{\widehat{\mathcal{N}}_{i}(t)}^{* \top}(\bm{x}_{\widehat{\mathcal{N}}_{i}(t)}^* - \bm{x}_{\widehat{\mathcal{N}}_{i}(t)}) - \bm{\theta}_{\widehat{\mathcal{N}}_{i}(t)}^{* \top}\bm{x}_{\widehat{\mathcal{N}}_{i}(t)}^* + \frac{1}{2} \bm{\theta}_i^{* \top}\bm{x}_i^* + \frac{1}{2} \bm{\theta}_j^{* \top}\bm{x}_j^*
\end{align*}

Moreover, since $\bm{\theta}_{\widehat{\mathcal{N}}_{i}(t)}^{* \top}\bm{x}_{\widehat{\mathcal{N}}_{i}(t)}^* \geq \bm{\theta}_{\widehat{\mathcal{N}}_{i}(t)}^{* \top}\left( \frac{1}{2}(\bm{x}_i^* + \bm{x}_j^*)\right)$, we have:
\begin{align*}
    \bar{r}_{i, t} &\leq \bm{\theta}_{\widehat{\mathcal{N}}_{i}(t)}^{* \top}(\bm{x}_{\widehat{\mathcal{N}}_{i}(t)}^* - \bm{x}_{\widehat{\mathcal{N}}_{i}(t)}) -\bm{\theta}_{\widehat{\mathcal{N}}_{i}(t)}^{* \top}\left( \frac{1}{2}(\bm{x}_i^* + \bm{x}_j^*)\right) + \frac{1}{2} \bm{\theta}_i^{* \top}\bm{x}_i^* + \frac{1}{2} \bm{\theta}_j^{* \top}\bm{x}_j^* \\
    &\leq \bm{\theta}_{\widehat{\mathcal{N}}_{i}(t)}^{* \top}(\bm{x}_{\widehat{\mathcal{N}}_{i}(t)}^* - \bm{x}_{\widehat{\mathcal{N}}_{i}(t)}) + \frac{1}{2} \left( -\frac{1}{2}\bm{\theta}_i^{* \top}\bm{x}_i^* - \frac{1}{2}\bm{\theta}_i^{* \top}\bm{x}_j^* -\frac{1}{2}\bm{\theta}_j^{* \top}\bm{x}_i^* - \frac{1}{2}\bm{\theta}_j^{* \top}\bm{x}_j^* + \bm{\theta}_i^{* \top}\bm{x}_i^* + \bm{\theta}_j^{* \top}\bm{x}_j^* \right) \\
    &\leq \bm{\theta}_{\widehat{\mathcal{N}}_{i}(t)}^{* \top}(\bm{x}_{\widehat{\mathcal{N}}_{i}(t)}^* - \bm{x}_{\widehat{\mathcal{N}}_{i}(t)}) + \frac{1}{4} (\bm{\theta}_i^* - \bm{\theta}_j^*)^\top (\bm{x}_i^* - \bm{x}_j^*) \\
    &\leq 2 \beta(\delta, \bm{A}_{\widehat{\mathcal{N}}_{i}(t)}) \|\bm{x}_{\widehat{\mathcal{N}}_{i}(t)}\|_{\bm{A}_{\widehat{\mathcal{N}}_{i}(t)}^{-1}} + \frac{1}{4} (\bm{\theta}_i^* - \bm{\theta}_j^*)^\top (\bm{x}_i^* - \bm{x}_j^*) \\
    &\leq \sqrt{2} \beta(\delta, 2\bm{A}_t) \|\bm{x}_{\widehat{\mathcal{N}}_{i}(t)}\|_{\bm{A}_t^{-1}} + \frac{1}{4} (\bm{\theta}_i^* - \bm{\theta}_j^*)^\top (\bm{x}_i^* - \bm{x}_j^*) \\
    &\leq 2 \left( \frac{1}{\sqrt{2}} \frac{\beta(\delta, 2\bm{A}_t)}{\beta(\delta, \bm{A}_t)} \right) \beta(\delta, \bm{A}_t) \|\bm{x}_{\widehat{\mathcal{N}}_{i}(t)}\|_{\bm{A}_t^{-1}} + \frac{1}{4} (\bm{\theta}_i^* - \bm{\theta}_j^*)^\top (\bm{x}_i^* - \bm{x}_j^*) \\    
    &\leq 2 \left( \frac{1}{\sqrt{2}} \sqrt{1 + \frac{d \log 2}{2\log \frac{1}{\delta}}} \right) \beta(\delta, \bm{A}_t) \|\bm{x}_{\widehat{\mathcal{N}}_{i}(t)}\|_{\bm{A}_t^{-1}} + \frac{1}{4} (\bm{\theta}_i^* - \bm{\theta}_j^*)^\top (\bm{x}_i^* - \bm{x}_j^*) \\ 
    &\leq 2 \mu'(\delta, d) \beta(\delta, \bm{A}_t) \|\bm{x}_{\widehat{\mathcal{N}}_{i}(t)}\|_{\bm{A}_t^{-1}} + \frac{1}{4} (\bm{\theta}_i^* - \bm{\theta}_j^*)^\top (\bm{x}_i^* - \bm{x}_j^*) \quad \text{with} \quad \mu'(\delta, d) = \frac{1}{\sqrt{2}} \sqrt{1 + \frac{d \log 2}{2\log \frac{1}{\delta}}} 
\end{align*}

With~\citet{Abbasi2011} and by setting $\nu(\delta, T_s) =  \sqrt{ 4 \beta(\delta, \bm{A}_{T_s})^2 T_s d \log (1 + \frac{T_s}{d} ) }$, we have:
\begin{equation*}
    \bar{R}_{0, T_s}^{\mathrm{collab}} \leq \mu'(\delta, d) \nu(\delta, T_s) + \frac{T_s}{4} (\bm{\theta}_i^* - \bm{\theta}_j^*)^\top (\bm{x}_i^* - \bm{x}_j^*)\enspace.
\end{equation*}

\end{proof}

Note that for reasonable values of $d$ and $\delta$, we have: $\mu'(\delta, d) \leq \mu(\delta, d, \gamma)$ which gives us the intuition that the network of agents is an improvement over the single agent case.

The same intuition applies for the second term $\frac{T_s}{4} (\bm{\theta}_i^* - \bm{\theta}_j^*)^\top (\bm{x}_i^* - \bm{x}_j^*)$. Since the bandit parameters and the arms norms are bounded, the only configuration for which this inner product is high, is when at least one arm is approximately in the direction of $\bm{\theta}_i^* - \bm{\theta}_j^*$. With this conditioning, the separation of the two bandit parameters is accelerated and the collaboration ceases shortly after.

In summary, taking into account the averaged cumulative pseudo regret better underlines the \emph{adaptive} aspect of our collaboration. 

\section{Appendix: Expected number of misassigned agents} 
\label{sec:supp_clustering_error}

We detail the proof of the upper-bound on the clustering error in~\autoref{th:neighboor_error} .

\thmf*

\begin{proof}[Proof of {\upshape\autoref{th:neighboor_error}}]

We have, at iteration $t>0$, $N_{i, t}^e = |\hat{N}_{i, t} - \rho_{m(i)} N|$ with $\hat{N}_{i, t}$ the number of agents within the estimated cluster, we have:
\begin{align*}
    \mathbf{E} \left[ N_{i, t}^e \right] &= \mathbf{E} \left[ |\hat{N}_{i, t} - \rho_{m(i)} N| \right] \\
     &= \mathbf{E} \left[ \hat{N}_{i, t} - \rho_{m(i)} N \right] \quad \text{from~\autoref{lm:claim}} \\
     &= \mathbf{E} \left[ \sum_{j=1}^{N} \bm{1}_{\{ \Psi(i, j, t) = 0 \}} - \rho_{m(i)} N \right] \\
     &= \sum_{j=1}^{N} \mathbf{P} \left[ \Psi(i, j, t) = 0 \right] - \rho_{m(i)} N
\end{align*}

Focusing on $ \mathbf{P} \left[ \Psi(i, j, t) = 0 \right]$, we have:
\begin{align*}
    \mathbf{P} \left[ \Psi(i, j, t) = 0 \right] &= \mathbf{P} \left[ \| \hat{\bm{\theta}}_i - \hat{\bm{\theta}}_j \|_{\bm{A}_t} \leq 2 \beta(\delta, \bm{A}_t) \right] \\
     & \leq \mathbf{P} \left[ \| \bm{Z}_{i, j, t} \|_{\bm{A}^{-1}_t} \geq 2 \beta(\delta, \bm{A}_t) - \| \bm{\theta}^*_i - \bm{\theta}^*_j \|_{\bm{A}_t} \right]
\end{align*}

By introducing $u > 0$ and by using the Chernoff bound, we have:
\begin{align*}
     \mathbf{P} \left[ \Psi(i, j, t) = 0 \right] &\leq \exp(-u) \quad \text{with} ~ \sqrt{2u + \log \mathrm{det}\bm{A}_t} = \| \bm{\theta}^*_i - \bm{\theta}^*_j \|_{\bm{A}_t}  - 2 \beta(\delta, \bm{A}_t) \\
     &\leq \exp \left( -\frac{1}{2} \left(\| \bm{\theta}^*_i - \bm{\theta}^*_j \|_{\bm{A}_t}  - 2 \beta(\delta, \bm{A}_t) \right)^2 + \frac{1}{2} \log \mathrm{det}\bm{A}_t \right) \\
    \mathbf{P} \left[ \Psi(i, j, t) = 0 \right] &\leq \sqrt{\frac{\mathrm{det}(\bm{A}_t)}{\mathrm{exp}( ( \Delta^{\min}_{\bm{\theta}^*, \bm{A}_t} - \gamma \beta(\delta, \bm{A}_t) )^2 )}}
\end{align*}

Which yield us the desired result:
\[
 \scalebox{1.04}{
    $\mathbb{E} \left[ N_i^e(t) \right] \leq  \left\lceil \left( \frac{N \sqrt{\mathrm{det}(\bm{A}_t)}}{\mathrm{exp}( \frac{1}{2} ( \Delta^{\min}_{\bm{\theta}^*, \bm{A}_t} - \gamma \beta(\delta, \bm{A}_t) )^2 )} - \rho_{m(i)} N \right) \right\rceil$
    }
\]
where $\Delta^{\min}_{\bm{A}_t} = \underset{i, j \quad i \neq j}{\min} \|\bm{\theta}_i^* - \bm{\theta}_j^*\|_{\bm{A}_t}$ and $\rho_{m(i)} N$ the number of agents in the cluster $m(i)$, the agent's cluster. \\

\end{proof}

\section{Appendix: Cumulative pseudo regret with clustered agents} 
\label{sec:supp_third_cumul_regret_ub}

\thmg*

\begin{proof}[Proof of {\upshape\autoref{th:upper_bound_clustering}}]

At round $t$, for a given agent $i$, by considering the optimistic arm $\bm{x}_{\widehat{\mathcal{N}}_{i}(t)}$ and the optimal arm $\bm{x}_{i}^*$, from Equation~\autoref{eq:instantaneous_regret_beginning}, we have:

\begin{align*}
    r_{i, t} &\leq (\tilde{\bm{\theta}}_{\widehat{\mathcal{N}}_{i}(t)} - \bm{\theta}_{\widehat{\mathcal{N}}_{i}(t)}^*)^\top\bm{x}_{\widehat{\mathcal{N}}_{i}(t)} +  (\bm{\theta}_i^* - \bm{\theta}^*_{\widehat{\mathcal{N}}_{i}(t)})^\top(\bm{x}_i^* - \bm{x}_{\widehat{\mathcal{N}}_{i}(t)}) \\
    &\leq (\tilde{\bm{\theta}}_{\widehat{\mathcal{N}}_{i}(t)} - \bm{\theta}^*_{\widehat{\mathcal{N}}_{i}(t)})^\top\bm{x}_{\widehat{\mathcal{N}}_{i}(t)} + 2 \|\bm{\theta}^*_{\widehat{\mathcal{N}}_{i}(t)} - \bm{\theta}_i^*\|_2 \quad \text{from~\autoref{assum:bounded_norms}} \\
    &\leq \left(\widehat{\bm{\theta}}_{\widehat{\mathcal{N}}_{i}(t)} + \frac{\beta(\delta, \bm{A}_{\widehat{\mathcal{N}}_{i}(t)})}{\|\bm{x}_{\widehat{\mathcal{N}}_{i}(t)}\|_{\bm{A}_{\widehat{\mathcal{N}}_{i}(t)}^{-1}}} \bm{A}_{\widehat{\mathcal{N}}_{i}(t)}^{-1 \top}\bm{x}_{\widehat{\mathcal{N}}_{i}(t)} - \bm{\theta}_{\widehat{\mathcal{N}}_{i}(t)}^*\right)^\top\bm{x}_{\widehat{\mathcal{N}}_{i}(t)} + 2 \|\bm{\theta}^*_{\widehat{\mathcal{N}}_{i}(t)} - \bm{\theta}_i^*\|_2 \quad \text{from~\autoref{lm:chara_theta_opt}} \\
    &\leq \beta(\delta, \bm{A}_{\widehat{\mathcal{N}}_{i}(t)}) \|\bm{x}_{\widehat{\mathcal{N}}_{i}(t)}\|_{\bm{A}_{\widehat{\mathcal{N}}_{i}(t)}^{-1}} + \left(\widehat{\bm{\theta}}_{\widehat{\mathcal{N}}_{i}(t)} - \bm{\theta}_{\widehat{\mathcal{N}}_{i}(t)}^*\right)^\top\bm{x}_{\widehat{\mathcal{N}}_{i}(t)} + 2 \|\bm{\theta}^*_{\widehat{\mathcal{N}}_{i}(t)} - \bm{\theta}_i^*\|_2 \\
    &\leq \left(\beta(\delta, \bm{A}_{\widehat{\mathcal{N}}_{i}(t)}) + \|\widehat{\bm{\theta}}_{\widehat{\mathcal{N}}_{i}(t)} - \bm{\theta}^*_{\widehat{\mathcal{N}}_{i}(t)}\|_{\bm{A}_{\widehat{\mathcal{N}}_{i}(t)}}\right) \|\bm{x}_{\widehat{\mathcal{N}}_{i}(t)}\|_{\bm{A}_{\widehat{\mathcal{N}}_{i}(t)}^{-1}} + 2 \|\bm{\theta}^*_{\widehat{\mathcal{N}}_{i}(t)} - \bm{\theta}_i^*\|_2 \\
    &\leq 2\beta(\delta, \bm{A}_{\widehat{\mathcal{N}}_{i}(t)})\|\bm{x}_{\widehat{\mathcal{N}}_{i}(t)}\|_{\bm{A}_{\widehat{\mathcal{N}}_{i}(t)}^{-1}} + 2 \|\bm{\theta}^*_{\widehat{\mathcal{N}}_{i}(t)} - \bm{\theta}_i^*\|_2 \\
    &\leq \frac{2}{\sqrt{N_i + N_i^{e}(t)}}\beta(\delta, (N_i + N_i^{e}(t)) \bm{A}_t)\|\bm{x}_{\widehat{\mathcal{N}}_{i}(t)}\|_{\bm{A}_t^{-1}} + 2 \|\bm{\theta}^*_{\widehat{\mathcal{N}}_{i}(t)} - \bm{\theta}_i^*\|_2 \\
    &\leq \frac{2}{\sqrt{N_i + N_i^{e}(t)}}\beta(\delta, (N_i + N_i^{e}(t)) \bm{A}_t)\|\bm{x}_{\widehat{\mathcal{N}}_{i}(t)}\|_{\bm{A}_t^{-1}} + 2  \left\| \frac{-N_i^{e}(t)}{N_i + N_i^{e}(t)} \bm{\theta}_i^* + \frac{1}{N_i + N_i^{e}(t)} \sum_{j \in \widehat{\mathcal{N}_i}\setminus\mathcal{N}_i}  \bm{\theta}_j^* \right\|_2 \\
    &\leq \frac{2}{\sqrt{N_i + N_i^{e}(t)}}\beta(\delta, (N_i + N_i^{e}(t)) \bm{A}_t)\|\bm{x}_{\widehat{\mathcal{N}}_{i}(t)}\|_{\bm{A}_t^{-1}} + 2  \frac{N_i^{e}(t)}{N_i + N_i^{e}(t)} L + \frac{1}{N_i + N_i^{e}(t)} \sum_{j \in \widehat{\mathcal{N}_i}\setminus\mathcal{N}_i} L \quad \text{from~\autoref{assum:bounded_norms}} \\
    &\leq \frac{2}{\sqrt{N_i + N_i^{e}(t)}}\beta(\delta, (N_i + N_i^{e}(t)) \bm{A}_t)\|\bm{x}_{\widehat{\mathcal{N}}_{i}(t)}\|_{\bm{A}_t^{-1}} + 2 \frac{2LN_i^{e}(t)}{N_i + N_i^{e}(t)} \\
    &\leq \frac{2}{\sqrt{N_i + N_i^{e}(t)}}\beta(\delta, (N_i + N_i^{e}(t)) \bm{A}_t)\|\bm{x}_{\widehat{\mathcal{N}}_{i}(t)}\|_{\bm{A}_t^{-1}} + \frac{4L}{\delta N_i}  n_i^e(t) \quad \text{from~\autoref{th:neighboor_error} and Markov inequality} \\
    &\leq \frac{2}{\sqrt{\rho_{\min}N}}\beta(\delta, N \bm{A}_t)\|\bm{x}_{\widehat{\mathcal{N}}_{i}(t)}\|_{\bm{A}_t^{-1}} + \frac{4L}{\delta \rho_{\min}N} n_i^e(t) \\
    &\leq 2 \left(\frac{1}{\sqrt{\rho_{\min}N}}\frac{\beta(\delta, N \bm{A}_t)}{\beta(\delta, \bm{A}_t)} \right)\beta(\delta, \bm{A}_t)\|\bm{x}_{\widehat{\mathcal{N}}_{i}(t)}\|_{\bm{A}_t^{-1}} + \frac{4L}{\delta \rho_{\min}N} n_i^e(t) \\
    &\leq 2 \left(\frac{1}{\sqrt{\rho_{\min}N}}\sqrt{1 + \frac{d \log N}{2\log \frac{1}{\delta}}} \right)\beta(\delta, \bm{A}_t)\|\bm{x}_{\widehat{\mathcal{N}}_{i}(t)}\|_{\bm{A}_t^{-1}} + \frac{4L}{\delta \rho_{\min}N} n_i^e(t) \\
    &\leq 2 \mu''(\delta, d, N) \beta(\delta, \bm{A}_t)\|\bm{x}_{\widehat{\mathcal{N}}_{i}(t)}\|_{\bm{A}_t^{-1}} + \frac{4L}{\delta \rho_{\min}N} n_i^e(t) \quad\text{with} \quad  \mu''(\delta, d, N) = \frac{1}{\sqrt{\rho_{\min}N}}\sqrt{1 + \frac{d \log N}{2\log \frac{1}{\delta}}}
\end{align*}

With~\citet{Abbasi2011} and by setting $\nu(\delta, T_s) =  \sqrt{ 4 \beta(\delta, \bm{A}_{T_s})^2 T_s d \log (1 + \frac{T_s}{d} ) }$, we have:
\begin{equation*}
    R_{i, 0, T}^{\mathrm{cluster}} \leq \mu''(\delta, d, N) \nu(\delta, T_s) + \frac{4L}{\delta \rho_{\min} N} C_i(T_c) \enspace,
\end{equation*}
with $C_i(T_c) = \sum_{t=1}^{\min(T, T_c)} n_i^e(t)$, $n_i^e(t)$ defined in~\autoref{th:neighboor_error}, $Tc = \max_j T_s(i, j)$ and $\rho_{\min} = \min_j \rho_{m(j)}$. 

Note that this upper bound is somewhat loose, as it introduces the iteration $T_c$ at which the cluster is adequately estimated. While it does not highlight the \emph{adaptive} nature of our collaboration, it still provides the asymptotic behavior of the \emph{BASS} algorithm.

\end{proof}

\section{Appendix: Additional theoretical analysis} 
\label{sec:supp_additional_analysis}

We gather additional theoretical results here to provide comparative baselines of our analysis.

\subsection{Proof of the cumulative pseudo regret upper-bound for the \emph{BASS-Oracle} algorithm}

\begin{restatable}[Cumulative pseudo regret upper-bound for \emph{Oracle}]{corollary}{corb}
    \label{cor:upper_bound_oracle}

    Under the assumptions made above, if we consider a \emph{Oracle} controller that know in advance the true clusters, after $T$ iterations the pseudo-regret can be bounded as follows:
    \vskip-1.5em
    \begin{equation*}
         R_{i, 0, T}^{\mathrm{oracle}} \leq \mu''(\delta, T) \nu(\delta, T) \enspace,
    \end{equation*}
    with $\mu''(\delta, d) = \frac{1}{\sqrt{N}} \sqrt{1 + \frac{d \log N}{2\log \frac{1}{\delta}}}$ and $\nu(\delta, T)$ defined as previously,

\end{restatable}

\begin{proof}[Proof of {\upshape\autoref{cor:upper_bound_oracle}}]

This is a direct application of~\autoref{th:upper_bound_clustering} with a null neighborhood estimation error \ie{} for $i$ and $t > 0$, we have $N_i^e(t) = 0$.

\end{proof}

\subsection{Asymptotic Analysis of Theoretical Bounds}
\label{sec:supp_asymptotic_analysis}

Building upon the result in~\autoref{th:upper_bound_clustering}, we derive the corresponding asymptotic behavior. First, we observe that the additive term in the bound is subdominant in \( T \) and can be ignored asymptotically. Furthermore, we have:
\[
\beta(\delta, \bm{A}_{T_s}) = \mathcal{O}\left( \sqrt{d \log T} \right) \enspace.
\]

Also, by comparing the confidence terms under collaborative and local designs, we obtain:
\[
\frac{1}{\sqrt{\rho_{\min}N}} \cdot \frac{\beta(\delta, N \bm{A}_t)}{\beta(\delta, \bm{A}_t)} = \mathcal{O}\left( \frac{\sqrt{d}}{\sqrt{\rho_{\min} N}} \right) \enspace,
\]
which leads to the following asymptotic regret bound:
\[
R_{i, 0, T}^{\mathrm{cluster}} = \mathcal{O}\left( \frac{\sqrt{d}}{\sqrt{\rho_{\min} N}} \cdot \sqrt{T} \cdot \log\left(\frac{T}{d}\right) \right) \enspace.
\]

Regarding the expected number of misassigned agents, we use the approximation:
\[
\sqrt{\det(\bm{A}_t)} = \mathcal{O}(t) \enspace,
\]
which implies the asymptotic expression:
\[
\mathcal{O}\left( \frac{N}{T^{\gamma^2 R^2 - 1/2}} \right) \enspace.
\]

This analysis, in conjunction with previous results, such as those of~\citet{Gentile2014,Li2019,Ghosh2022}, supports the comparison table.

\subsection{Summary of Clustering Assumptions}
\label{sec:supp_assumptions}

We summarize below the main clustering assumptions considered in the literature, as well as those adopted in our work:
\begin{itemize}[itemsep=0.1pt, topsep=0.15pt]
    \item \textit{No assumption:} considered in our paper, except for the setting of~\autoref{th:upper_bound_clustering}.
    \item \textit{Identical bandit parameters within each cluster:} used in our paper under~\autoref{assum:clustering_struct}.
    \item \textit{Identical bandit parameters within each cluster, with a minimum separation between clusters:} commonly assumed in prior work such as~\citet{Gentile2014,Li2016,Li2018,Ghosh2022,Wang2023,Yang2024}.
    \item \textit{Approximately similar parameters within clusters, allowing overlapping cluster memberships:} considered in~\citet{Ban2021}.
\end{itemize}

\section{Appendix: Implementation details} 
\label{sec:supp_implementation_details}

At each iteration, our \emph{BASS} algorithm calls at most $N-1$ time the function $\Psi$, which involves to inverse multiple matrices. We propose an efficient implementation involving the use of the Cholesky decomposition, the Sherman-Morrison identity or the eigenvalues decomposition. We report here the implementation details for the main parts of the \emph{BASS} algorithm and their associated complexity.

\paragraph{Agent separation test} Recall that the ellipsoid separation test relies on the function defined in~\autoref{def:ell_test}, which involves computing:
\begin{equation*}
    \forall s \in ]0, 1[ \quad \psi_{i,j}(s) = \frac{\gamma^2}{4} - \sum_{l=1}^d \mu_l^2 \frac{s (1 - s)}{\beta_i^2 + s (\beta_j^2\eta_l - \beta_i^2)} \enspace,
\end{equation*}
with $\bm{\mu} = \bm{\Phi}^\top \bm{v}$ such as $\bm{\eta}$ and $\bm{\Phi}$ are the eigenvalues and the eigenvectors of the matrix
\begin{equation*}
    \bm{Q} = \frac{\beta_i}{\beta_j} \text{chol}(\bm{A_j})^\top \bm{A_i}^{-1} \text{chol}(\bm{A_j}) \enspace,
\end{equation*}    
with $\text{chol}()$ being the Cholesky decomposition. Moreover, one can derive the first and the second derivatives of $\psi_{i,j}$, and performs the minimization of $\psi_{i, j}$ with the Newton's method. The eigen decomposition features a complexity of order $\mathcal{O}\left(d^3\right)$ and the computation of the first and second derivatives has a complexity of order $O\left(d\right)$. Hence, the complexity of the separation test is of order $\mathcal{O}\left(\tau d + d^3\right)$, with $\tau$ being the number of iterations of the minimization algorithm. However since we use a second order optimization method, in practice only a couple of iterations is needed (\ie{} $\tau < 50$), so practically the separation test is of order $\mathcal{O}\left(d^3\right)$.

\paragraph{Arm selection} Recall that the arm selection strategy, at iteration $t$, for an agent $i_t$ is defined as $k_t = \underset{k=1, \ldots, K}{\argmax} \quad \thetaNithat^\top \bm{x_k} + \beta_{\ANithat^{-1}}(\bm{x_k}) \quad \text{ with } \beta_{\ANithat^{-1}}(\bm{x}) = \alpha \sqrt{\bm{x}^\top \ANithat^{-1} \bm{x} \log(t)}$ which features a complexity of $\mathcal{O}\left(Kd^2\right)$. \\

Following the different implementations detailed here, the overall complexity of the \emph{BASS} algorithm is of order $\mathcal{O}\left( T(N + Kd^2 + (N - 1)d^3) \right)$. We notice that the \emph{BASS} algorithm as a linear complexity toward the number of agents $N$, the number of arms $K$ and the number of iterations $T$. However, it features a cubic dependence toward the dimension $d$ of the problem.

\section{Appendix: Experimental details} 
\label{sec:supp_experiment_details}

\paragraph{Description of the concurrent algorithms} \quad  For the benchmarks, we select four other concurrent algorithms: \emph{DynUCB}, \emph{CLUB}, \emph{SCLUB}, \emph{CMLB}. First, \emph{DynUCB} is a simple yet efficient approach to perform agent clustering in a linear bandit setting. Proposed in~\citet{Nguyen2014}, it relies on the $k$-means algorithms~\citet{MacQueen1967} to update the clusters at each iteration. This algorithm needs to set the number of estimated clusters. For this experiment, we set it to the true number of clusters. To our knowledge, no other method proposes this $k$-means-based approach for this class of problems, which makes it of particular interest. Second, \emph{CLUB} and its improved version \emph{SCLUB} are likely the most benchmarked algorithms for the clustering linear bandit problem and can be considered the reference for it. \emph{CLUB} was introduced in~\citet{Gentile2014} and \emph{SCLUB} resp. in~\citet{Li2019}, both propose to determine whether two agents belong to the same cluster by performing a test on the $\ell_2$-norm of the difference of their bandit parameters. The improved version \emph{SCLUB} provides a more flexible cluster estimation by iteratively splitting and merging the clusters. Finally, we consider \emph{CMLB} form~\citet{Ghosh2022} which is a more recent approach closely related to \emph{CLUB} using the same method to determine whether two agents belong to the same cluster. However, this approach controls the minimum number of agents in a cluster. Note that these last three algorithms need to set a scalar to scale the clustering threshold that determines whether this difference is significant.

\paragraph{Synthetic experiment settings} \quad  In~\autoref{sec:exp}, in the synthetic experiment, we consider $M=3$ clusters of the same size with bandit parameter $(\bm{\theta}_m)_{m \in \{1, \ldots, M\}}$ defined as: $\forall 1 \leq q \leq \lceil \frac{M}{2} ~ \rceil \bm{\theta}_{2q-1} = \bm{e}_q$, with $(\bm{e}_i)_i$ being the canonical basis and $\bm{\theta}_{2q}$ having its $q$-th entry being $\cos{\omega}$, its $(q+1)$-th entry being $\sin{\omega}$ and all the other entries being $0$, with $\omega \in \{\pi/16, 7\pi/16\}$. Indeed, these configurations are of particular interest because they correspond to the case where the bandit parameters are positively correlated and almost orthogonal. With $\omega = \pi/16$, the bandit parameters $(\bm{\theta}_m)_{m \in \{1, \ldots,M\}}$ are similar. Thus, assigning the wrong cluster will not overly perturb the agent, even though discriminating clusters will be tougher. The second scenario with $\omega = 7\pi/16$ the two bandit parameters are almost orthogonal which makes them easier to separate one another and thus identify the cluster. However, the two models are very different thus if an agent is assigned to the wrong cluster, he will most likely perform poorly once he will gather the observations of the other agents. For the \emph{CLUB}, \emph{SCLUB} and \emph{CMLB} algorithms we test two levels of clustering exploration $\gamma \in \{\frac{\Delta^{\mathrm{min}}_{\bm{\theta}^*}}{4}, \frac{\Delta^{\mathrm{min}}_{\bm{\theta}^*}}{2}\}$, the result is shown in the `dashed' and `solid' lines. For \emph{DynUCB}, we explore $M=5$ and $M=20$ for the $k$-means algorithms~\citet{MacQueen1967} with the same visual coding for the two cases. For the \emph{BASS} algorithm, we test two values for $\delta \in \{0.1, 0.9\}$ equivalently, with the same choice of result visualization. We keep the same color code for the rest of the experiment: \emph{DynUCB} in pink, \emph{CLUB} in blue, \emph{SCLUB} in brown, \emph{CMLB} in red and our approach \emph{BASS} in orange, \emph{Ind} in light gray and when available \emph{Oracle} in dark gray. We choose the $\alpha$-\emph{LinUCB} agent policy for our algorithm and line-search the UCB parameter $\alpha$ within $[0.1, 3.0]$.

\paragraph{Clustering score computation} \quad To quantify the degree of agreement between the estimated clusters/neighborhoods and the ground truth, we compute the $\mathrm{F}_1$-score on the graph $\mathcal{G}_t$ w.r.t. the true cluster graph $\mathcal{G}$. For the \emph{DynUCB} we define $\mathcal{G}_t = (V, E_t)$ such as $E_t = \{i,j| 1_{\{\mathrm{kmeans-label}(i) = \mathrm{kmeans-label}(j)\}}\}$. The clustering score will output a value within $[0, 1]$, where $1$ is a perfect cluster match.

\paragraph{Description of the datasets and experiment settings} \quad Recall that in~\autoref{sec:exp}, we perform experiments on two real datasets. We detail here their characteristics. First we select the \emph{MovieLens} dataset\footnote{Available at \url{https://grouplens.org/datasets/movielens/}}~\citet{Harper2015}: which is a dataset that proposes the rating (from $1$ to $5$) of 62,000 movies by 162,000 users for a total of 25 million ratings. Each movie is represented by features describing its genre and historical records. Since the users did not rate all the movies systematically, we fill the missing values with a matrix-factorization approach~\citet{Rendle2008}. From this complete dataset, we are able, for a given user---agent---and a given movie---arm---, to fetch the corresponding reward.

Second, we consider, \emph{Yahoo!} dataset\footnote{Available at \url{https://webscope.sandbox.yahoo.com/catalog.php?datatype=c}}~\citet{Chapelle2011}: which is a dataset of recommendations made from queries on the Yahoo! search engine. For each query, documents are received, represented by sparse features vectors of dimension $500$ and their corresponding reward (from $1$ to $4$) which measures how relevant the returned documents were. We consider the queries as---agents---and the document as the---arms---, to reduce their cardinality, we perform the $k$-means algorithm and retained only $K=30$ arms. As with the previous dataset, we complete the preprocessing by filling the missing rewards with the same matrix-factorization approach to obtain, for each user, a reward for every arm.

We consider the same setting as the synthetic experiment: we select $N=100$ agents and reduce the arm dimension to $d=10$ by taking the $d$ first dimensions of the native data space. Since the randomness of our preprocessing is the user's draft to be retained on the $N$ agents, we run the experiment $50$ times to average across runs.

\section{Appendix: Additional details} 
\label{sec:supp_additional_experiments}

\subsection{Appendix: Additional synthetic experiment results}

To complete the analysis with the synthetic data scenario considered in~\autoref{sec:exp}, we report here the evolution of the cumulative regret value in the case of $M=3$ and $M=6$. We consider the same experimental setting as in the synthetic data experiment of~\autoref{sec:exp}. \\

\begin{figure}[H]
    \centering
    \setlength{\tabcolsep}{5pt}
    \begin{tabular}{>{\centering\arraybackslash}m{1.5cm}>{\centering\arraybackslash}m{7cm}>{\centering\arraybackslash}m{7cm}}
        & \textbf{\large (a)} & \textbf{\large (b)} \\
        \textbf{\large M=3} & \includegraphics[width=.40\textwidth,trim= 0 160 0 0,clip]{figures/figure_regret_evolution_Rt_synth_data__n_thetas_3.pdf} &
        \includegraphics[width=.4\textwidth]{figures/figure_regret_evolution_wrt_alpha_synth_data__n_thetas_3.pdf} \\
        \textbf{\large M=6} & \includegraphics[width=.40\textwidth,trim= 0 160 0 0,clip]{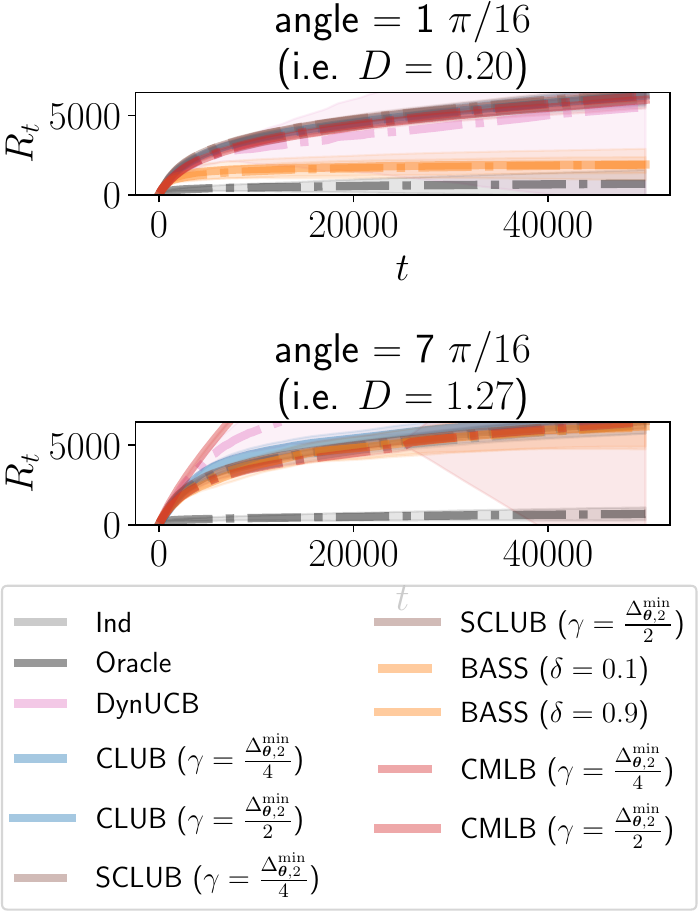} &
        \includegraphics[width=.4\textwidth]{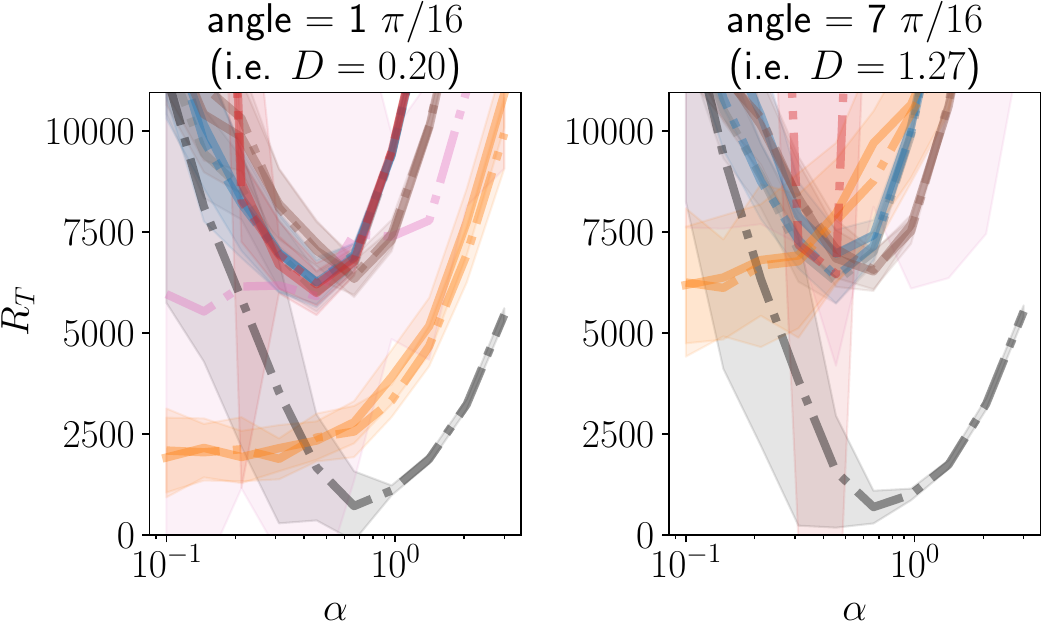} \\
    \end{tabular}
    \caption{\textbf{(a)} Comparison of the averaged, across runs, cumulative regret evolution $(R_t)_t$ for the synthetic environments considered, $M=3$ on the top and $M=6$ on the bottom. \textbf{(b)} Comparison of the averaged, across runs, evolution of the cumulative regret last value $R_T$ \wrt{} the UCB parameter $\alpha$ for the different synthetic environments considered, with the same color code as previously.}
    \label{fig:figure_regret_evolution_Rt_synth_data__n_thetas_3_and_6}
\end{figure}
\vskip-.5em

In~\autoref{fig:figure_regret_evolution_Rt_synth_data__n_thetas_3_and_6} \textbf{a} we display the cumulative regret evolution $(R_t)_t$, $M=3$ on the top and $M=6$ on the bottom and in~\autoref{fig:figure_regret_evolution_Rt_synth_data__n_thetas_3_and_6} \textbf{b}, we display the cumulative regret last value, $R_T$, evolution w.r.t the UCB parameter $\alpha$, $M=3$ on the top and $M=6$ on the bottom, both for all the concurrent and baseline methods. The overall performance comparison between the algorithms stays coherent with all the previous experiments and confirms the robustness and the good behavior of our algorithm. Interestingly, for \emph{BASS}, we systematically observe an inflection of the curve to a quasi no-regret plateau. \\

\subsection{Appendix: Additional real data experiment results}

To complete the analysis with the real data scenario considered in the paper, we report here the evolution of the cumulative regret last value \wrt{} the UCB parameter.  

\begin{figure}[H]
    \centering
    \includegraphics[width=.50\textwidth]{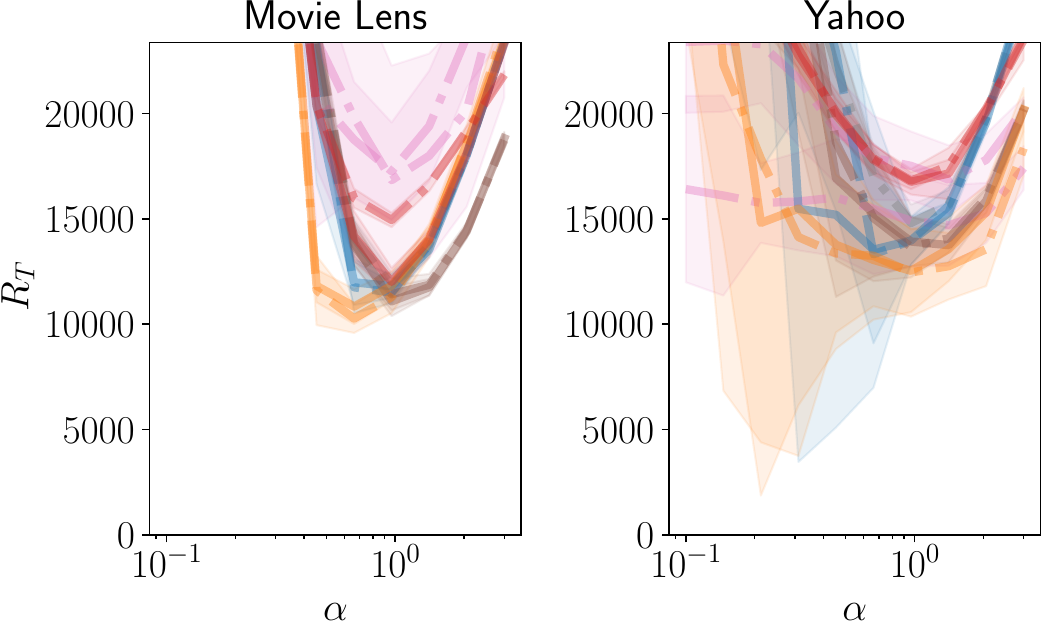}
    \vskip-.75em
    \caption{Comparison of the averaged, across runs, evolution of the cumulative regret last value $R_T$ \wrt{} the UCB parameter $\alpha$ for the different synthetic environments considered, with the same color code as previously.
    \label{fig:figure_regret_evolution_wrt_alpha_real_data}}
\end{figure}
\vskip-.5em

In \autoref{fig:figure_regret_evolution_wrt_alpha_real_data}, we display the cumulative regret last value, $R_T$, evolution \wrt{} the UCB parameter $\alpha$. We notice that again \emph{BASS} performs better than the other algorithms, it is more pronounced in the case of the \emph{Yahoo} dataset. 

\subsection{Appendix: Additional benchmark experiment}

Additionally, we consider a simpler benchmark to underline the robustness of the results depicted in the paper. In this benchmark, we consider the same experimental setting as in the synthetic data experiment of ~\autoref{sec:exp}, but in this case, we randomly draw the bandit parameters and the arms from a standard Gaussian distribution and consider $M=3$. We consider three levels of noise with $\sigma \in \{0.5, 1.0, 2.0\} $

\begin{figure}[H]
    \centering
    \setlength{\tabcolsep}{5pt}
    \begin{tabular}{>{\centering\arraybackslash}m{7cm}>{\centering\arraybackslash}m{7cm}}
        \includegraphics[width=.35\textwidth,trim= 0 160 0 0,clip]{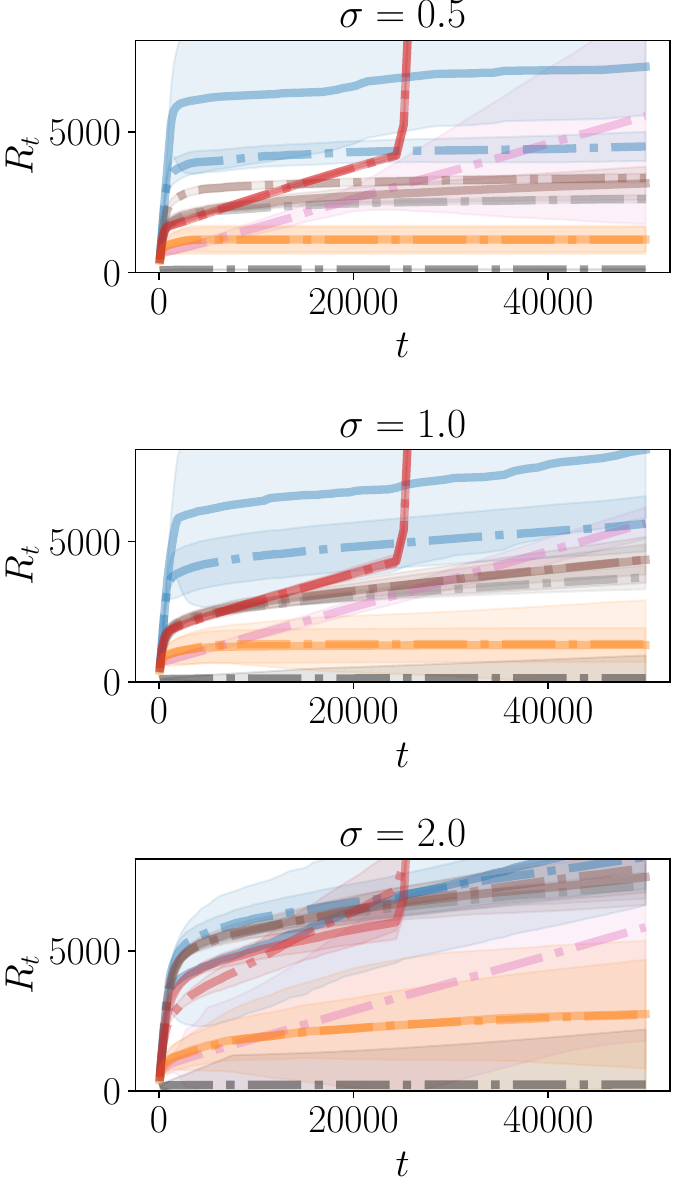} &
        \includegraphics[width=0.45\linewidth,trim= 2 2 2 2,clip]{figures/figure_regret_evolution_wrt_alpha_synth_data__n_thetas_3_legend.pdf} \\
    \end{tabular}
    \vskip-.75em
    \caption{Comparison of the averaged, across runs, cumulative regret evolution $(R_t)_t$ for the different synthetic environments considered with $M=3$.
     \label{fig:supp_figure_regret_evolution_Rt_synth_data__n_thetas_3_and_6}}
\end{figure}

In \autoref{fig:supp_figure_regret_evolution_Rt_synth_data__n_thetas_3_and_6}, we display the cumulative regret evolution $(R_t)_t$ for all the concurrent and baseline methods. We notice the good performance of the \emph{BASS} algorithm and interestingly in the case where $\sigma=2.0$, we notice that the difference of performance compare to the other increase, which emphasize how robust to the noise level our algorithm is.
    
\begin{figure}[H]
    \centering
    \begin{tabular}{>{\centering\arraybackslash}m{7cm}>{\centering\arraybackslash}m{7cm}}
        \includegraphics[width=.35\textwidth,trim= 0 160 0 0,clip]{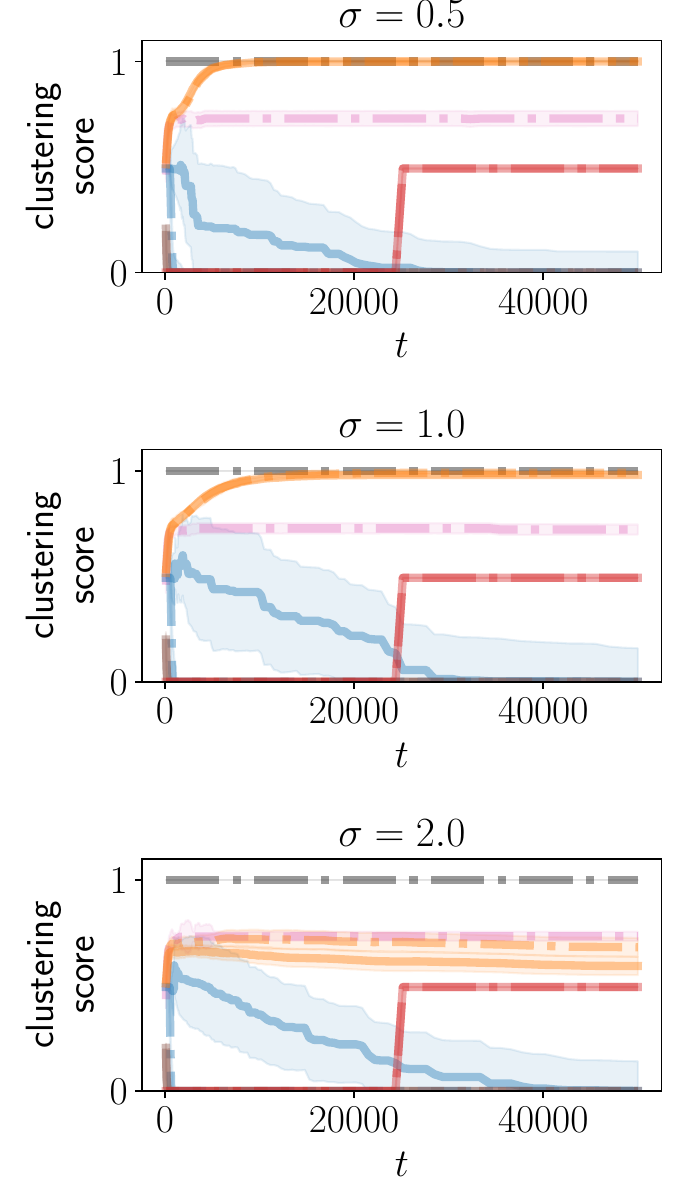} &
        \includegraphics[width=0.45\linewidth,trim= 2 2 2 2,clip]{figures/figure_regret_evolution_wrt_alpha_synth_data__n_thetas_3_legend.pdf} \\
    \end{tabular}
    \vskip-.75em
    \caption{Comparison of the averaged, across runs, clustering score evolution for the different synthetic environments considered with $M=3$.
    \label{fig:supp_figure_clustering_score_evolution_synth_data__n_thetas_3_and_6}}
\end{figure}
\vskip-.5em

In \autoref{fig:supp_figure_clustering_score_evolution_synth_data__n_thetas_3_and_6}, we display the clustering score evolution for all the clustering algorithms. We quantify the clustering estimation quality as previously and notice that our approach achieves again the best performance. Indeed, we observe that most of the concurrent algorithms did not manage recover any clusters.

\end{document}